\documentclass[10pt,twocolumn,letterpaper]{article}

\usepackage[pagenumbers]{cvpr} 

\usepackage{graphicx}
\usepackage{amsmath}
\usepackage{amssymb}
\usepackage{booktabs, multirow}
\usepackage{rotating}
\usepackage[labelfont=bf,skip=2pt]{caption}
\usepackage{enumitem}
\usepackage[accsupp]{axessibility}
\usepackage{titling}

\usepackage[dvipsnames]{xcolor}
\usepackage{float}
\definecolor{OliveGreen}{rgb}{0,0.6,0}
\definecolor{SoftRed}{rgb}{1,0.2,0.2}
\usepackage[pagebackref,breaklinks,colorlinks,citecolor=OliveGreen,linkcolor=SoftRed]{hyperref}

\usepackage[capitalize]{cleveref}
\crefname{section}{Sec.}{Secs.}
\Crefname{section}{Section}{Sections}
\Crefname{table}{Table}{Tables}
\crefname{table}{Tab.}{Tabs.}

\def\cvprPaperID{8783} 
\def\confName{CVPR}
\def\confYear{2022}

\newcommand{\rotateT}[1]{\begin{turn}{90}\footnotesize{#1}\end{turn}}
\newcommand{\ts}[1]{\scriptsize{#1}}
\newcommand{\tsb}[1]{\scriptsize{\textbf{#1}}}
\newcommand{\FLOATsys}{FLOAT}
\predate{}
\postdate{}
\date{}

\title{\textbf{FLOAT: Factorized Learning of Object Attributes for Improved Multi-object Multi-part Scene Parsing}}
\author{Rishubh Singh\textsuperscript{1} \hspace{1cm} Pranav Gupta\textsuperscript{2} \hspace{1cm} Pradeep Shenoy\textsuperscript{1} \hspace{1cm} Ravikiran Sarvadevabhatla\textsuperscript{2}\\
\textsuperscript{1}Google Research \hspace{1cm} \textsuperscript{2} IIIT Hyderabad\\
{\tt\small \{rishubh,shenoypradeep\}@google.com, \{ravi.kiran@,pranav.gu@research.\}iiit.ac.in}}

\begin{document}
\twocolumn[{%
\renewcommand\twocolumn[1][]{#1}%
\maketitle
\begin{center}
    \centering
    \captionsetup{type=figure}
    \includegraphics[width=\textwidth]{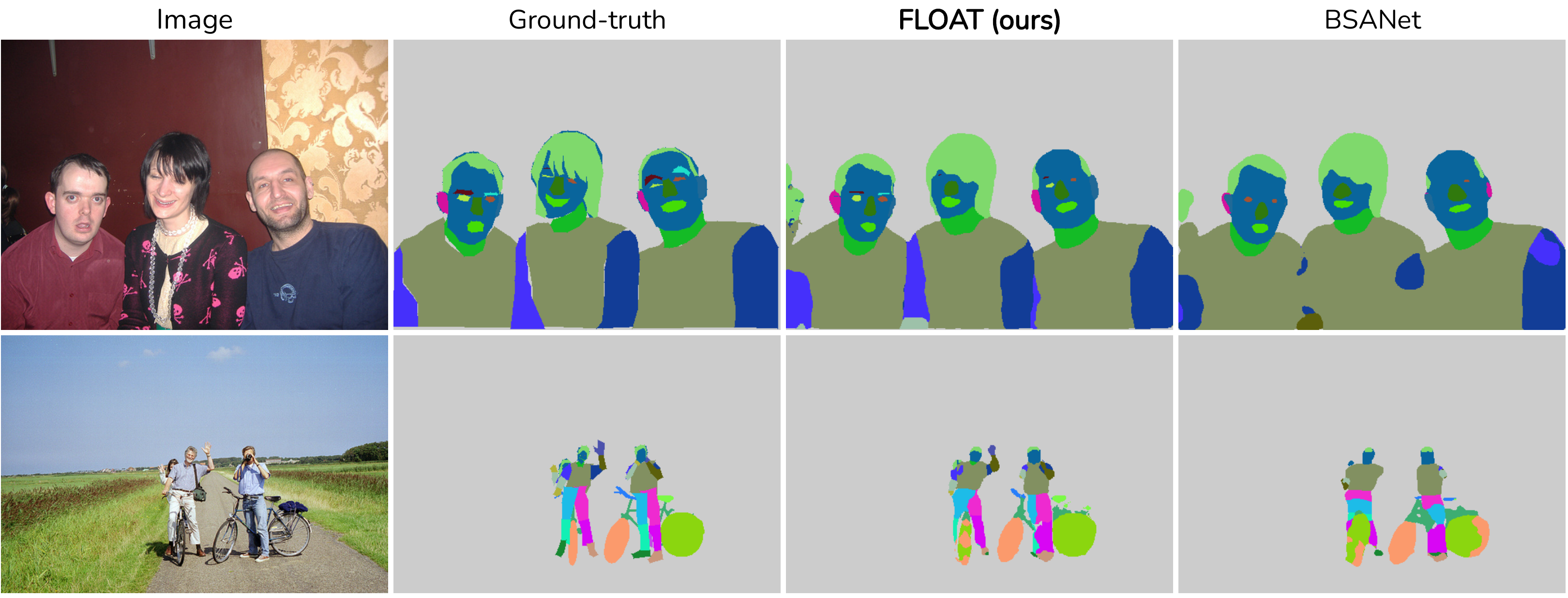}
    \captionof{figure}{Multi-object multi-part semantic segmentation results for sample images from our expanded label space dataset, Pascal-Part-201. Compared to state of the art BSANet~\cite{bsanet}, FLOAT accurately segments tiny parts (e.g. \texttt{left} \texttt{eyebrow}, \texttt{right} \texttt{eyebrow} on faces in upper image) and handles scale variations better -- note the size variations of \texttt{person} instances. Also, observe that FLOAT predicts directional attributes of parts (e.g. `left'/`right') accurately -- [`left'/`right']: see \texttt{eyebrow}, \texttt{eye}, \texttt{arm} in upper image and \texttt{leg}  in lower image ; [`front'/`back']: see \texttt{wheel} parts of the \texttt{bicycle} (lower image).}
    \label{fig:fig1}
\end{center}
}]

\begin{abstract}
\vspace{-6pt}
    Multi-object multi-part scene parsing is a challenging  task which requires detecting multiple object classes in a scene and segmenting the semantic parts within each object. In this paper, we propose FLOAT, a factorized label space framework for scalable multi-object multi-part parsing. Our framework involves independent dense prediction of object category and part attributes which increases scalability and reduces task complexity compared to the monolithic label space counterpart. In addition, we propose an inference-time `zoom' refinement technique which significantly improves segmentation quality, especially for smaller objects/parts. Compared to state of the art, FLOAT obtains an absolute improvement of 2.0\% for mean IOU (mIOU) and 4.8\% for segmentation quality IOU (sqIOU) on the Pascal-Part-58 dataset. For the larger Pascal-Part-108 dataset, the improvements are 2.1\% for mIOU and 3.9\% for sqIOU. We incorporate previously excluded part attributes and other minor parts of the Pascal-Part dataset to create the most comprehensive and challenging version which we dub Pascal-Part-201. FLOAT obtains improvements of 8.6\% for mIOU and 7.5\% for sqIOU on the new dataset, demonstrating its parsing effectiveness across a challenging diversity of objects and parts. The code and datasets are available at \href{https://floatseg.github.io/}{floatseg.github.io}.
\end{abstract}

\vspace{-12pt}

\section{Introduction}
\label{sec:intro}

Semantic scene parsing is a foundational image understanding problem in the vision community~\cite{zheng2021rethinking, zhao2018icnet, li2020improving, yu2018bisenet, yang2018denseaspp, zhang2018exfuse, yuan2020object}. Typically, the goal is to segment objects and ``stuff" regions (e.g. road, background) in the scene. Multi-object multi-part parsing is a significantly more challenging variant which requires \textit{part}-level segmentation of each scene object~\cite{bsanet, gmnet, co-rank}. Compared to traditional object-level segmentation, semantic representations infused with fine-grained part-level knowledge can provide richer information for downstream reasoning tasks including visual question answering~\cite{hong2021ptr}, perceptual concept learning~\cite{DBLP:journals/corr/abs-2111-05251}, shape modelling~\cite{achlioptas2019shapeglot,dubrovina2019composite} and many others ~\cite{dong2014humanparsing,chen2014detect,10.1007/978-3-642-33718-5_60,DBLP:journals/corr/ZhangDGD14, sun2013learning,krause2015fine}.

For part-based object segmentation, some existing approaches tackle the simpler problem of \textit{single}-object part parsing~\cite{gong2018instance, fang2018weakly, wang2015joint, wang2015semantic, haggag2016semantic}. Although a few recent approaches have addressed multi-object multi-part parsing~\cite{bsanet, gmnet, co-rank}, they consider part labels to be independent and do not take advantage of intra/inter ontological relationships among objects and parts at label level. They also tend to perform poorly on smaller and infrequent parts/categories. To address these shortcomings, we propose \FLOATsys, a novel factorized label space framework for scalable multi-object multi-part parsing.  Our approach is motivated by the following observations:

\noindent \textbf{Observation \#1:} Object part names in datasets typically consist of a {\color{RedOrange} \textit{root}} component and {\color{ForestGreen} \textit{side}} component(s). Many object categories contain parts with the same {\color{RedOrange} \textit{root}} component. For example, the {\color{RedOrange} \textit{root}} component of `{\color{ForestGreen} left} {\color{OliveGreen} front} {\color{RedOrange} leg}' found in \texttt{horse}, \texttt{cow} etc. and `{\color{ForestGreen} right} {\color{RedOrange} leg}' found in \texttt{person}, is {\color{RedOrange} leg}. Therefore, parts can be grouped based on their {\color{RedOrange} \textit{root}} component. 

The example also suggests that object categories whose instances contain shared category-level attributes (e.g. ``living things that move") are likely to contain same {\color{RedOrange} \textit{root}} components (such as {\color{RedOrange} leg}). Using this criterion, some object categories (e.g. \texttt{cow, person, bird}) can be grouped as `animate'. Similarly, some categories (e.g. ``rigid bodied") can be grouped as `inanimate'. As with the `animate' group, `inanimate' group categories also share many {\color{RedOrange} \textit{root}} part components (e.g. `{\color{RedOrange}wheel}'  in \texttt{aeroplane}, \texttt{bicycle}, \texttt{car}). 

 \noindent \textbf{Observation \#2:} Similar to Observation \#1, parts can also be grouped by  {\color{ForestGreen} \textit{side}} component -- e.g. `{\color{OliveGreen} front}' is a {\color{ForestGreen} \textit{side}} component of `{\color{OliveGreen} front} {\color{RedOrange}wheel}' found in \texttt{bike} and `{\color{ForestGreen} left} {\color{OliveGreen} front} {\color{RedOrange} leg}' in \texttt{person}.

Factoring the object/part label space in terms of these groups (`animate', `inanimate', `side') greatly reduces the effective number of output labels. In turn, this increases scalability in terms of object categories and part cardinality. The  design choice (`factoring') also enables efficient data sharing when learning semantic representations for grouped parts and improves performance for infrequent classes (see Fig.~\ref{fig:fig1}). 

A second key feature of our framework is IZR, an \textit{inference-time} segmentation refinement technique. IZR transforms `zoomed in' versions of preliminary per-object label maps into refined counterparts which are finally composited back onto the segmentation canvas. Apart from the advantage of not requiring additional training, IZR is empirically superior to alternate inference-time schemes and significantly improves segmentation quality, especially for smaller objects/parts. 

In existing works, results are reported on simplified, label-merged versions of the original dataset (Pascal-Part~\cite{chen2014detect}). In our work, we incorporate previously excluded part attributes and other minor parts to create Pascal-Part-201, the most comprehensive and challenging version of  Pascal-Part~\cite{chen2014detect}. Along with the standard mean IOU (mIOU) and mAvg scores, we report sqIOU~\cite{kirillov2019panoptic} and sqAvg -- normalized segmentation quality measures which are less affected by spatial scale of objects and parts.

In summary, our contributions are the following:
\begin{itemize}[noitemsep]
    \item \FLOATsys, a novel factorized label space framework for scalable multi-object multi-part parsing (Sec.~\ref{sec:float}).
    \item IZR, an inference-time refinement technique which significantly improves segmentation quality especially for smaller objects/parts in the scene (Sec.~\ref{sec:izr}).
    \item Pascal-Part-201, the most comprehensive and challenging version of the Pascal-Part~\cite{chen2014detect} dataset (Sec.~\ref{sec:datasetsnmetrics}). Experimental evaluation demonstrates \FLOATsys's superior performance on Pascal-Part-201 relative to existing approaches (Sec.~\ref{sec:experiments}).
\end{itemize}

\section{Related Work}
\label{sec:related}

\textbf{Semantic segmentation} is a broad area with intensive research. We do not attempt to summarize all approaches to enable focus on more directly relevant works. A common design pattern for semantic segmentation is the encoder-decoder setup~\cite{7803544,zhao2017pyramid,chen2017deeplab,article_123}. In particular, the baselines, existing approaches and our proposed approach all adopt the popular DeepLab architecture~\cite{chen2017deeplab} for various components of the segmentation task pipeline.

\textbf{Single-Object Multi-Part Parsing} has been extensively explored. Existing approaches typically consider object category subsets such as persons~\cite{fang2018weakly, liang2018look, liang2016semantic, nie2018mutual, xia2016zoom, xia2017joint, xia2015pose, zhao2017self, gong2018instance, liang2015human, luo2018macro, liu2020hybrid}, animals~\cite{haggag2016semantic, wang2015semantic, wang2015joint} and vehicles~\cite{liang2016semantic, nie2018mutual, song2017embedding, liu2021cgpart}. However, in this setting, most works assume a single object of interest per image.

\begin{figure*}[!t]
  \centering
    \includegraphics[width=0.95\linewidth]{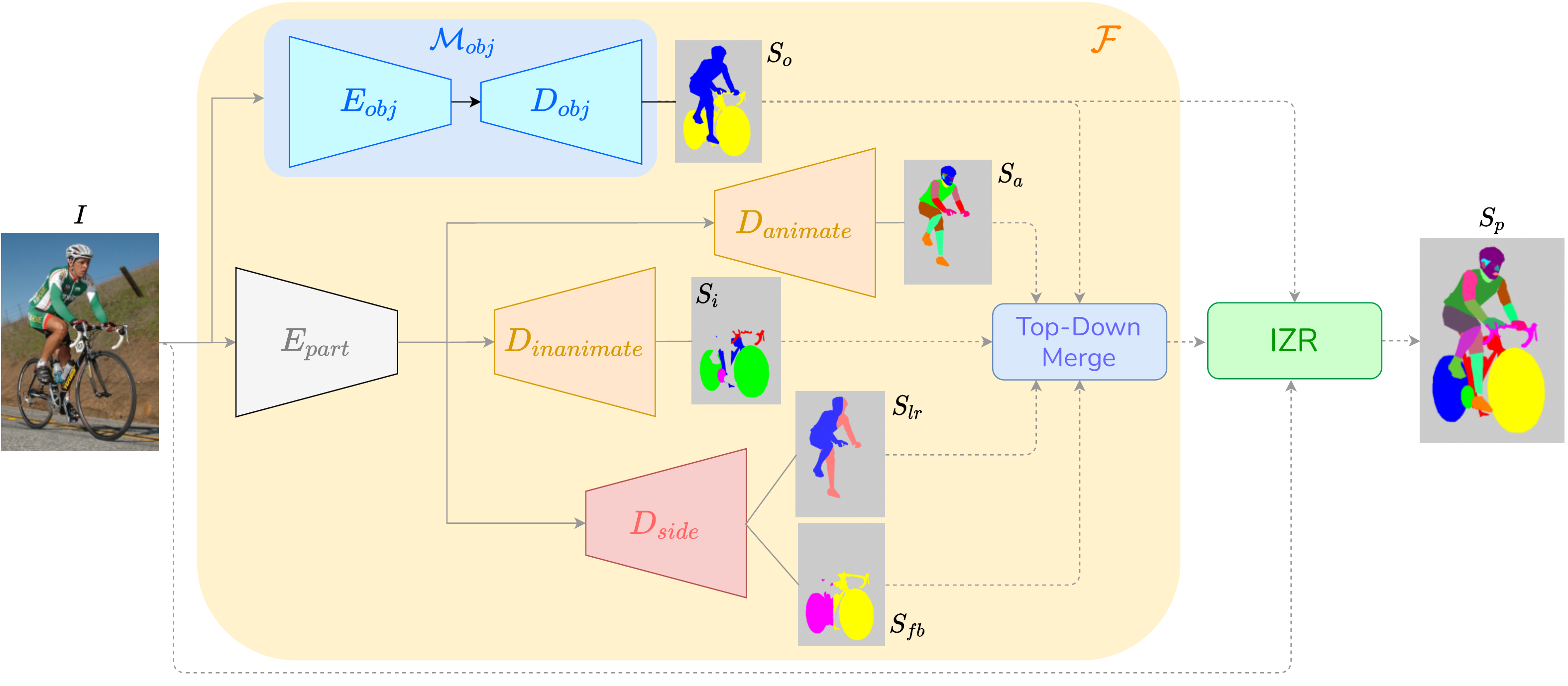}
    \caption{An overview diagram of our FLOAT framework (Sec.~\ref{sec:float}). Given an input image $I$, an object-level semantic segmentation network ($\mathcal{M}_{obj}$, in blue) generates object prediction map ($S_o$). Two  decoders (in orange) produce object category grouped part-level prediction maps for `animate' ($S_a$) and `inanimate' objects ($S_i$) in the scene. Another decoder (in red) produces part-attribute grouped prediction maps for `left-right' ($S_{lr}$) and `front-back' ($S_{fb}$). At inference time (shown by dotted lines), outputs from the decoders are merged in a top-down manner. The resulting prediction is further refined using the IZR technique (see Fig.~\ref{fig:zoom}) to obtain the final segmentation map ($S_p$).}
    \label{fig:float}
\end{figure*}

\textbf{Multi-object multi-part parsing} is a relatively new and under studied problem~\cite{bsanet, gmnet, co-rank}. The approaches of Zhao et al.~\cite{bsanet} and Michieli et al.~\cite{gmnet} tackle multi-object multi-part parsing by providing object-level feature guidance to the part segmentation network during optimization. Zhao et al.~\cite{bsanet} additionally provides boundary-level awareness to features. Tan et al.~\cite{co-rank} create a semantic co-ranking loss modelling intra and inter part relationships. Xiao et al.~\cite{xiao2018unified} introduce a composite dataset and an approach for predicting perceptual visual concepts in scenes. However, in contrast to our framework, these approaches report results on simplified (label-merged) versions of standard datasets and empirically exhibit inferior performance for smaller parts. 

\textbf{Factorization:} In machine vision applications, early works such as Zheng et al.~\cite{DenseObjAtt_CVPR2014} used factorial Conditional Random Field models to  separately predict object category, coarse object labels and object attributes such as shape, material and surface type. Other works involve jointly learning object and attribute-related information as a separable latent representation~\cite{nagarajan2018attributes} or using graph networks~\cite{naeem2021learning}. Misra et al.~\cite{misra2017red} propose a factorization over global object attributes and object classifiers to enable compositionality. Other works extend this idea to inter-object relationships, e.g. noun-preposition-noun triplets~\cite{malinowski2014pooling,lan2012image,hong2021ptr}. In all these works, a simple \textit{global} property of the object (e.g., material, texture, color, size, shape) is learnt jointly with the object category information. In their work on panoptic part segmentation, Geus et al.~\cite{de2021part} conduct experiments involving two categories from Pascal-Part-58 with some parts grouped by semantic similarity.  Graphonomy, a framework by Lin et al. ~\cite{lin2020graphonomy} can span multiple datasets with a flat label structure and requires a manually specified graph per category. Such rigid connectivity relationships are unsuitable for modelling highly articulated objects (e.g. animals) found in our setting. To the best of our knowledge, we are the first to show that \textit{object parts} can be factorized across diverse object categories at scale, and that such factorization significantly improves segmentation performance, in resonance with theories of visual recognition~\cite{biederman1987recognition,HOFFMAN198465}.

\textbf{Zooming in} on image regions using bounding boxes generated by attention maps~\cite{wang2017zoom} and reinforcement learning policies~\cite{dong2018reinforced, xu2021adazoom} have been found to improve detection and segmentation. Other works use the technique on object instances for video interpolation~\cite{yuan2019zoom-in-to-check} and on part instances for object parsing~\cite{xia2016zoom}. Porzi et al.~\cite{porzi2021improving} use zoomed in crops based on object classes for improving panoptic segmentation of high resolution images. Similar to the latter set of approaches, \FLOATsys\ also employs zooming in on object regions. However, our zoom-based refinement does not require any extra training and can be directly used during inference for improved performance.

\section{Our framework (FLOAT)}
\label{sec:float}

As mentioned earlier, FLOAT's design leverages the shared-attribute groups that naturally exist within object categories (`animate', `inanimate') and part attributes (`left', `right', `front', `back') - see Fig.~\ref{fig:float}. The sections that follow describe how we operationalize the idea. Although our approach is general in nature, we use object categories and part names from the Pascal-Part dataset~\cite{chen2014detect} for ease of understanding.

\begin{figure*}[!t]
  \centering
    \includegraphics[width=\linewidth]{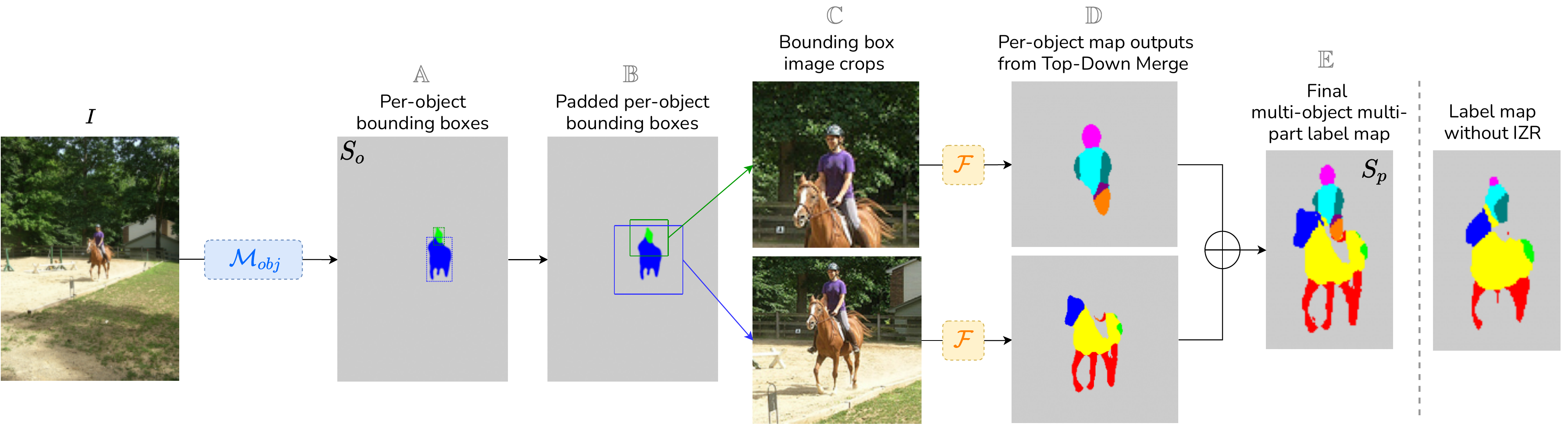}
    \caption{An overview of  Inference-time Zoom Refinement (IZR) - Sec.~\ref{sec:izr}. During inference, predictions from the object-level network $\mathcal{M}_{obj}$ are used to obtain padded bounding boxes for scene objects ($\mathbb{B}$). The corresponding object crops ($\mathbb{C}$) are processed by the factorized network ($\mathcal{F}$, Sec.~\ref{sec:float}). The resulting label maps ($\mathbb{D}$) are composited to generate $S_p$, the final refined part segmentation map ($\mathbb{E}$). Notice the improvement in segmentation quality relative to the part label map without IZR (included for comparison).}
    \label{fig:zoom}
    \vspace{-2mm}
\end{figure*}

\subsection{Relabeling images with factored labels}
\label{sec:relabeling}

The original Pascal-Part dataset contains object and part level label maps. We re-label or partition these maps to obtain five new label groups as described below. 

{\color{ProcessBlue}\textbf{object}:} The label set for this group comprises unique object category labels. For example, $S_o$ in Fig.~\ref{fig:float} is a label map from this group containing \texttt{person} and \texttt{bicycle} objects.

{\color{RedOrange}\textbf{animate}:} For this group, the label set comprises \textit{root} components of part labels from the object categories \texttt{bird}, \texttt{cat}, \texttt{cow}, \texttt{cat}, \texttt{dog}, \texttt{horse}, \texttt{person}, \texttt{sheep}. The part labels are pooled across all object categories. For example, a single label {\color{RedOrange} leg} covers all corresponding part instances from all objects in the `animate' group. This can also be seen in $S_a$ in Fig.~\ref{fig:float} -- the \texttt{left {\color{RedOrange} foot}} and \texttt{right {\color{RedOrange} foot}} of \texttt{person} are color-coded the same (`orange') and assigned the common label \texttt{{\color{RedOrange} foot}}. 

{\color{RedOrange} \textbf{inanimate}:} The label set comprises \textit{root} components of part labels from  \texttt{aeroplane}, \texttt{bicycle}, \texttt{bottle}, \texttt{bus}, \texttt{car},  \texttt{motorbike}, \texttt{pottedplant}, \texttt{train}, \texttt{tv}. Note that (i) these categories are disjoint from the `animate' group (see $S_i$ in Fig.~\ref{fig:float}) (ii) the part label pooling mentioned for `animate' is applicable here as well.

{\color{SoftRed}\textbf{side}:} In this case, two disjoint label groups exist. One group comprises all part labels which have the words `{\color{SoftRed}left}' or `{\color{SoftRed}right}' in their name (e.g. {\color{SoftRed}\texttt{left}} \texttt{hand}, {\color{SoftRed}\texttt{right}} \texttt{wing}). Label map regions whose part labels contain `{\color{SoftRed}left}'/`{\color{SoftRed}right}' are considered seed pixels for a flood-fill style procedure which produces corresponding `{\color{SoftRed}left}'/`{\color{SoftRed}right}' label maps (e.g. $S_{lr}$ in Fig.~\ref{fig:float}). The same procedure is used for the label groups which have the words `{\color{SoftRed}front}' or `{\color{SoftRed}back}' in their name (see $S_{fb}$ in Fig.~\ref{fig:float}). {\color{ProcessBlue}Appendix A.2} contains detailed explanation of the flood-fill algorithm.

Broadly, object parts from living things that move are in the `animate' group while other parts, typically from rigidly shaped non-living things, are in the `inanimate' group. As mentioned before, such grouping enables data-efficient representation learning for common parts (e.g. \texttt{torso} in `animate' group). A similar reasoning holds for `side' directional grouping (\{`left', `right'\}, \{`front',`back'\}). 

\subsection{Factorized semantic segmentation architecture}
\label{sec:factoredseg}

We configure the segmentation architecture to output the factorized label maps described in previous section. As Fig.~\ref{fig:float} shows, we employ two semantic segmentation networks, one for object-level and other for part-level label maps. The object-level network ($\mathcal{M}_{obj}$) outputs the object prediction map ($S_o$). The part-level network consists of a shared encoder ($E_{part}$), and three decoders: the `animate' decoder ($D_{animate}$) which outputs the `animate' label map ($S_a$), the `inanimate' decoder ($D_{inanimate}$) which outputs the `inanimate' label map ($S_i$). The `side' decoder ($D_{side}$) outputs the `left/right' ($S_{lr}$) and `front/back' ($S_{fb}$) label maps. The outputs from the object-level network ($S_o$) and part-level network ($S_i, S_a, S_{lr}, S_{fb}$) are merged at inference time. We describe this merging process next.

\subsection{Top-Down Merge}
\label{sec:topdown-merge}

To combine the factorized label maps output by segmentation architecture $\mathcal{F}$ (see Fig.~\ref{fig:float}), we adopt a top-down merging strategy. For each object (e.g. \texttt{bicycle}) in the object prediction map ($S_o$), we examine the labels of corresponding pixel locations in the part-level label maps. Depending on the type of object (`animate' or `inanimate'), the corresponding label regions are copied to the scene-level prediction canvas. (e.g. for \texttt{bicycle}, the considered labels in $S_i$ would be \texttt{wheel}, \texttt{chainwheel},  \texttt{handlebar}, \texttt{headlight}, \texttt{saddle}). Similarly, the object-level map's pixel locations are referenced from `side' label maps (\{`left',`right'\} - $S_{lr}$, \{`front',`back'\} - $S_{fb}$). In case of conflicts, the prediction defaults to \texttt{background}. The corresponding label regions are copied to the scene prediction canvas. Detailed explanation of top-down merging can be found in {\color{ProcessBlue}Appendix A.1} .

In the next section, we describe how the resulting prediction map is refined using a per-object `zooming' technique.

\subsection{Inference-time Zoom Refinement (IZR)}
\label{sec:izr}

The Inference-time Zoom Refinement (IZR) technique improves segmentation quality by `zooming' into each scene object. As the first step, the input image $I$ is processed by the object-level network $\mathcal{M}_{obj}$ to obtain object-level map (see $\mathbb{A}$ in Fig.~\ref{fig:zoom}). The bounding box corresponding to each object component is then padded so that the object is centered and aspect ratio is preserved ($\mathbb{B}$ in Fig.~\ref{fig:zoom}). Image crops corresponding to the padded bounding box extents are then obtained ($\mathbb{C}$). Note that the padding enables scene context to be included for each cropped object and also helps account for inaccuracies in the object map prediction. The cropped object images are then processed by FLOAT's factorized network $\mathcal{F}$ to obtain the corresponding part-level label maps ($\mathbb{D}$). These label maps are then composited to generate the final refined segmentation map ($\mathbb{E}$). In the next two sections, we describe the optimizer formulation for the networks in \FLOATsys\ and implementation details.

\subsection{Optimization}
\label{sec:optimization}

We train the object model $\mathcal{M}_{obj}$ (Sec.~\ref{sec:factoredseg}) using the standard per-pixel cross-entropy loss. For training the part-level model, we use a combination of cross-entropy loss ($L_{CE}$) and graph matching loss ($L_{GM}$)~\cite{gmnet}. The cross-entropy loss is applied to each of the 4 output part-level maps i.e. $S_a, S_i, S_{lr}, S_{fb}$ (see Fig.~\ref{fig:float}). 

The graph matching loss~\cite{gmnet} captures proximity relationships between part pairs within the map and scores the matching of these pairs between the ground truth and the predicted map. The degree of proximity between a part pair is represented by the number of pixels in one part situated $T$ pixels or less from the other part, where $T$ is an empirically set threshold. For efficiency, the pairwise proximity map is approximated by dilating each part mask by $\lceil T/2 \rceil$ and computing the intersecting region. The ground truth proximity map $M^{GT}$ (and similarly predicted map $M^{pred}$) is formally defined as: $\tilde{m}_{i,j}^{GT} = |\{s\in \Phi(p_i^{GT}) \cap \Phi(p_j^{GT})\}|$ where $\tilde{m}_{i,j}^{GT}$ is the proximity between the $i$th and $j$th parts, $p_i, p_j$ are the respective part mask, $s$ is a generic pixel, $\Phi$ is morphological 2D dilation operator and $|.|$ is the cardinality of the given set. A row-wise normalization is applied to the proximity matrix: $\pmb{M}_{[i,:]}^{GT} = \pmb{\tilde{M}}_{[i,:]}^{GT} / ||\pmb{\tilde{M}}_{[i,:]}^{GT}||_2$. The graph matching loss $L_{GM}$ is computed as the Frobenius norm between the two adjacency matrices: $\mathcal{L}_{GM} = ||\pmb{M}^{GT} - \pmb{M}^{pred}||_F$. 

Additionally, for the `animate' and `inanimate' branches, a composite foreground-background binary cross-entropy loss serves as extra guidance. The loss for the part level network is a weighted combination of the losses for all part branches: $\mathcal{L}^{part} =  \mathcal{L}^{anim} + \mathcal{L}^{inanim} + \mathcal{L}^{side}$, where $\mathcal{L}^{anim} = \mathcal{L}_{CE}^{anim} + \lambda_{GM}\mathcal{L}_{GM}^{anim}$.

\subsection{Implementation and Training Details}
\label{sec:impl}

For fair comparison with previous works~\cite{bsanet, gmnet,co-rank}, we employ the DeepLab-v3~\cite{chen2017deeplab} architecture with a ImageNet pre-trained ResNet-101~\cite{he2016deep} as the encoder (backbone) and follow the same training scheme and augmentations. During training, images are randomly left-right flipped and scaled $0.5$ to $2$ times the original resolution with bilinear interpolation. The results at testing stage are reported at the original image resolution. The threshold $T$ employed for proximity matrix (Sec.~\ref{sec:optimization}) is empirically set to $4$. The model is trained for 40K steps with the base learning rate set to $7 \cdot 10^{-3}$ which is decreased with a polynomial decay rule with power $0.9$. We employ weight decay regularization of $10^{-4}$. We use a batch size of $16$ images and use $\lambda_{GM} = 0.1$ for weighting graph matching loss relative to the cross-entropy loss. We use 2 NVIDIA A100 GPUs each with 40GB GPU memory to train our models, and for experiments. Full computational and memory requirement can be found in {\color{ProcessBlue}Appendix C}. 

\section{Datasets and Evaluation Metrics}
\label{sec:datasetsnmetrics}

\begin{figure}
  \centering
    \includegraphics[width=\linewidth]{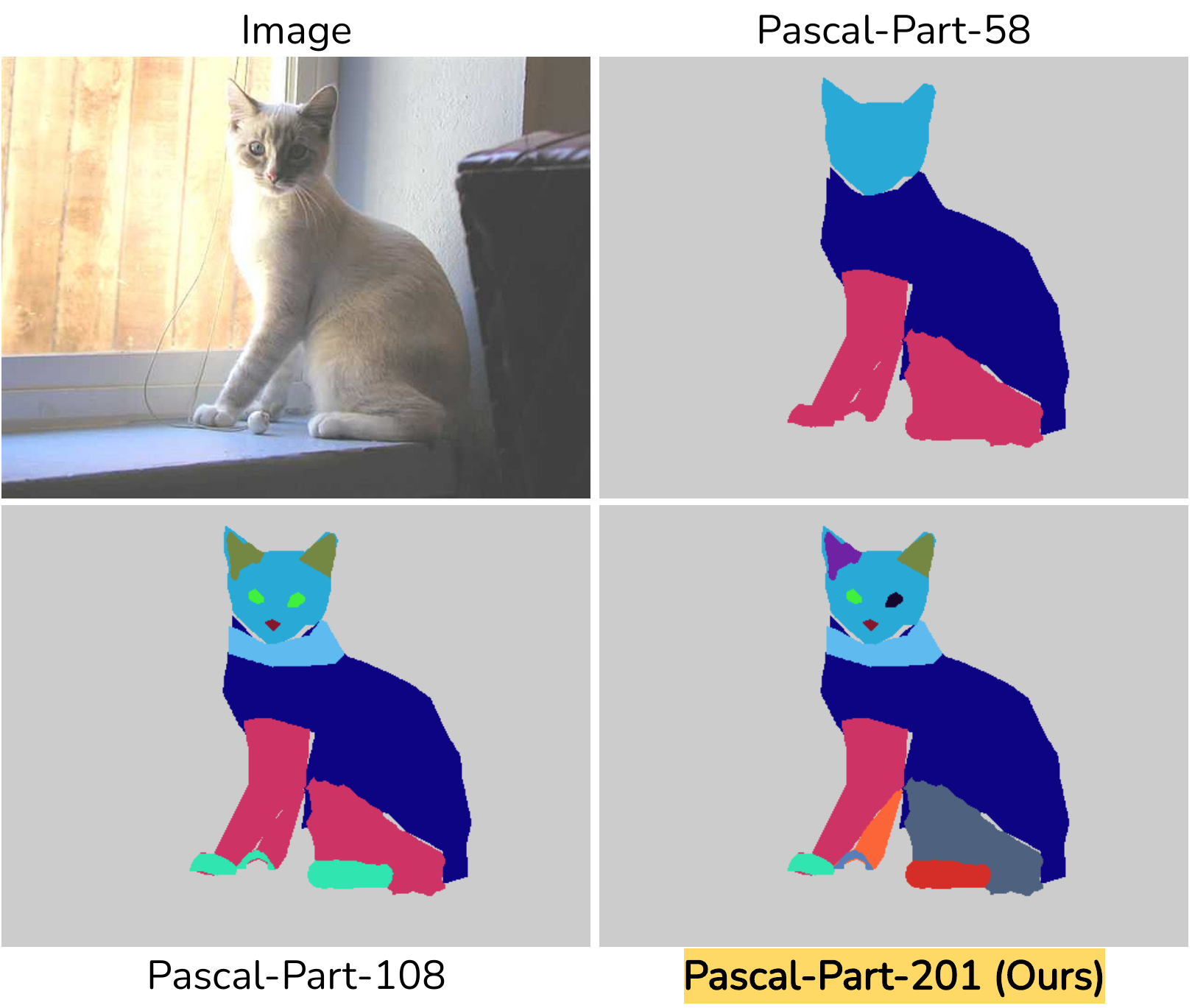}
    \caption{An illustration of labelling granularity in different versions of the Pascal-Part dataset. Pascal-Part-108~\cite{gmnet} adds smaller parts (e.g. \texttt{eyes}, \texttt{ears}) to Pascal-Part-58~\cite{bsanet}. Our newly introduced Pascal-Part-201 further adds directional information to parts as appropriate  (e.g. \{`left',`right'\} to \texttt{eyes}, \texttt{ears}; \{`front',`back'\} to \texttt{legs}).}
    \label{fig:pascal-part}
\end{figure}

\noindent \textbf{Pascal-Part:} For experiments, we use the Pascal-Part~\cite{chen2014detect} which is currently the largest multi-object multi-part parsing dataset. It contains $10{,}103$ variable-sized images with pixel-level part annotations on the $20$ Pascal VOC2010~\cite{everingham2010pascal} semantic object classes (plus the \texttt{background} class). We use the original split from Pascal-Part with $4998$ images for training and $5105$ images in the publicly provided validation set for testing.
 
 \noindent \textbf{Pascal-Part-58/108:} For comparison with previous work, we use the datasets Pascal-Part-58~\cite{bsanet} and Pascal-Part-108~\cite{gmnet} which contain $58$ and $108$ part classes respectively. Both the Pascal-Part variants simplify the original semantic classes by grouping some parts together, and contain $58$ and $108$ part classes respectively.  Pascal-Part-58 mostly contains large parts of objects such as \texttt{head}, \texttt{torso}, \texttt{leg} etc. for animals and \texttt{body}, \texttt{wheel} etc. for non-living objects. Pascal-Part-108 is more challenging and additionally contains relatively smaller parts (e.g. \texttt{eye}, \texttt{neck}, \texttt{foot} etc. for animals and \texttt{roof}, \texttt{door} etc. for non-living objects).
 
 \noindent \textbf{Pascal-Part-201:}
 We incorporate part attributes (`left', `right', `front', `back', `upper', `lower') and other minor parts (e.g. \texttt{eyebrow}) excluded in both the mentioned variants (58/108), to create the most comprehensive and challenging version of the dataset containing $201$ parts which we dub Pascal-Part-201. We observed that the original part labelling scheme in Pascal-Part leaves out large chunks of an object's pixels unlabelled for the \texttt{bike}, \texttt{motorbike} and \texttt{tv} categories which lead to disconnected objects. To address this, we add a \texttt{body} part annotation for \texttt{bike}, \texttt{motorbike}, and a \texttt{frame} part for \texttt{tv}. An example illustrating the differences in part labelling and granularity of the Pascal-Part variants can be seen in Fig.~\ref{fig:pascal-part}.

\begin{table*}[!t]
  \centering
  \setlength\tabcolsep{3.4pt}
  \begin{tabular}{@{}l|ccccccccccccccccccccc|cc@{}}
    \toprule
    \footnotesize{Model} &  \rotateT{bgr} & \rotateT{aero} & \rotateT{bike} & \rotateT{bird} & \rotateT{boat} & \rotateT{bottle} & \rotateT{bus} & \rotateT{car} & \rotateT{cat} & \rotateT{chair} & \rotateT{cow} & \rotateT{table} & \rotateT{dog} & \rotateT{horse} & \rotateT{mbike} & \rotateT{person} & \rotateT{plant} & \rotateT{sheep} & \rotateT{sofa} & \rotateT{train} & \rotateT{TV} & \ts{\textbf{mIOU}} & \ts{\textbf{mAvg}} \\
    \midrule
    \footnotesize{Baseline} & \ts{91.0} & \ts{31.6} & \ts{47.7} & \ts{24.3} & \ts{56.7} & \ts{46.4} & \ts{31.0} & \ts{36.7} & \ts{24.2} & \ts{35.6} & \ts{17.5} & \ts{38.6} & \ts{27.3} & \ts{20.7} & \ts{38.0} & \ts{26.9} & \ts{50.8} & \ts{13.3} & \ts{42.1} & \ts{14.7} & \ts{57.6} & \ts{26.3} & \ts{36.8} \\
    \footnotesize{GMNet\cite{gmnet}}  & \ts{90.8} & \ts{26.6} & \ts{33.1} & \ts{21.2} & \ts{55.0} & \ts{43.5} & \ts{24.6} & \ts{27.5} & \ts{21.7} & \ts{35.5} & \ts{15.1} & \ts{40.3} & \ts{25.0} & \ts{17.5} & \ts{31.9} & \ts{21.9} & \ts{44.2} & \ts{11.9} & \ts{43.3} & \ts{14.0} & \ts{53.2} & \ts{22.5} & \ts{33.2}\\
    \footnotesize{BSANet\cite{bsanet}} & \ts{91.2} & \ts{34.6} & \ts{41.7} & \ts{27.9} & \ts{61.2} & \ts{51.7} & \ts{34.1} & \ts{38.1} & \ts{26.1} & \ts{35.4} & \ts{24.0} & \ts{43.6} & \ts{28.4} & \ts{23.0} & \ts{37.4} & \ts{27.7} & \tsb{54.7} & \ts{14.3} & \ts{40.4} & \tsb{17.8} & \tsb{59.4} & \ts{28.5} & \ts{38.7}\\
    \midrule
    \footnotesize{\textbf{FLOAT}} & \tsb{92.5} & \tsb{36.7} & \tsb{49.7} & \tsb{34.4} & \tsb{75.3} & \tsb{51.4} & \tsb{35.8} & \tsb{42.0} & \tsb{37.8} & \tsb{59.6} & \tsb{35.5} & \tsb{58.2} & \tsb{41.0} & \tsb{34.0} & \tsb{40.2} & \tsb{40.8} & \ts{52.2} & \tsb{28.5} & \tsb{69.0} & \ts{15.1} & \ts{56.1} & \tsb{37.1} & \tsb{46.9} \\
    \bottomrule
    \toprule
     &  \rotateT{bgr} & \rotateT{aero} & \rotateT{bike} & \rotateT{bird} & \rotateT{boat} & \rotateT{bottle} & \rotateT{bus} & \rotateT{car} & \rotateT{cat} & \rotateT{chair} & \rotateT{cow} & \rotateT{table} & \rotateT{dog} & \rotateT{horse} & \rotateT{mbike} & \rotateT{person} & \rotateT{plant} & \rotateT{sheep} & \rotateT{sofa} & \rotateT{train} & \rotateT{TV} & \ts{\textbf{sqIOU}} & \ts{\textbf{sqAvg}} \\
    \midrule
    \footnotesize{Baseline} & \ts{89.6} & \ts{28.9} & \ts{39.3} & \ts{17.1} & \ts{57.4} & \ts{32.3} & \ts{27.1} & \ts{26.0} & \ts{20.5} & \ts{39.8} & \ts{14.8} & \ts{34.7} & \ts{22.7} & \ts{17.2} & \ts{31.5} & \ts{19.2} & \ts{34.9} & \ts{10.8} & \ts{52.6} & \ts{14.4} & \ts{53.8} & \ts{21.5} & \ts{32.6} \\
    \footnotesize{GMNet\cite{gmnet}}  & \ts{89.4} & \ts{20.7} & \ts{23.5} & \ts{12.6} & \ts{53.1} & \ts{25.8} & \ts{19.3} & \ts{17.2} & \ts{18.1} & \ts{38.2} & \ts{11.2} & \ts{35.2} & \ts{15.9} & \ts{14.2} & \ts{25.4} & \ts{13.8} & \ts{26.9} & \ts{8.5} & \ts{52.0} & \ts{13.8} & \ts{46.9} & \ts{16.9} & \ts{27.7}\\
    \footnotesize{BSANet\cite{bsanet}}  & \ts{89.9} & \ts{30.7} & \ts{33.5} & \ts{18.6} & \ts{60.2} & \ts{31.2} & \ts{29.2} & \ts{26.4} & \ts{21.2} & \ts{37.8} & \ts{17.5} & \ts{38.0} & \ts{22.3} & \ts{17.8} & \ts{31.2} & \ts{18.2} & \ts{33.6} & \ts{10.8} & \ts{47.2} & \tsb{17.5} & \ts{55.4} & \ts{22.1} & \ts{32.8}\\
    \midrule
    \footnotesize{\textbf{FLOAT}} & \tsb{90.8} & \tsb{32.5} & \tsb{41.8} & \tsb{24.5} & \tsb{63.9} & \tsb{36.1} & \tsb{30.4} & \tsb{29.9} & \tsb{33.0} & \tsb{50.8} & \tsb{28.1} & \tsb{47.6} & \tsb{35.6} & \tsb{26.1} & \tsb{33.6} & \tsb{29.9} & \tsb{34.5} & \tsb{20.6} & \tsb{69.0} & \ts{13.6} & \tsb{56.8} & \tsb{29.6} & \tsb{39.5} \\
    \bottomrule
  \end{tabular}
  \caption{Category-wise results for Pascal-Part-201. FLOAT outperforms competing methods by large margins w.r.t mIOU (top) and sqIOU (bottom).
  }
  \label{tab:results201}
  \vspace{-3mm}
\end{table*}

\subsection{Evaluation Metrics}
\label{sec:metrics}

For performance evaluation, we use two versions of Intersection over Union (IOU) metric. We first describe mIOU and mAvg, the standard segmentation quality metrics reported for the problem setting. We then describe balanced variants of these metrics -- sqIOU and sqAvg.

\noindent \textbf{mIOU:} Let $Pred_p^j$ and $GT_p^j$ be the prediction and ground truth respectively for the $p$th part in the $j$th image $I_j$. Suppose the dataset contains $N$ images.  The mIOU for the part ($mIOU_p$) is calculated as:

\begin{equation}
    mIOU_p = \dfrac{\sum_{j = 1}^{N} (Pred_p^j \cap GT_p^j)\cdot \mathbb{I}[p \in I_j]}{\sum_{j = 1}^{N} (Pred_p^j \cup GT_p^j)\cdot \mathbb{I}[p \in I_j]}
    \label{eqn:miou-p}
\end{equation}

where $\mathbb{I}[.]$ is the indicator function (i.e. summation is performed only for images where part $p$ is present). The mIOU for the dataset is then calculated as: $mIOU = \left(\sum_p mIOU_p\right) / N_p$, where $N_p$ is the number of part categories (classes) in the dataset (58/108/201).

\noindent \textbf{mAvg:} The mIOU score for an object category is the average of its per-part scores, i.e. $mIOU_c = \left(\sum_p mIOU_p\right) / N_c$ where $N_c$ is the number of unique part labels in object category $c$. Finally, mAvg is calculated as  $\text{mAvg} = \left(\sum_c mIOU_c\right) / C$, where $C$ is the number of object categories ($21$ for Pascal-Part datasets).

\noindent \textbf{sqIOU:} This is a modified version of Segmentation Quality (SQ) metric~\cite{kirillov2019panoptic} tailored for semantic segmentation. The sqIOU for the part $p$ is calculated as:

\begin{equation}
    sqIOU_p = \sum_{j = 1}^{N} \left(\underbrace{\dfrac{Pred_p^j \cap GT_p^j}{Pred_p^j \cup GT_p^j}}_{IOU(Pred,GT)_p^j} \cdot \mathbb{I}[p \in I_j]\right)/ N
    \label{eqn:sqiou-p}
\end{equation}

\begin{figure}
  \centering
  \includegraphics[width=\linewidth]{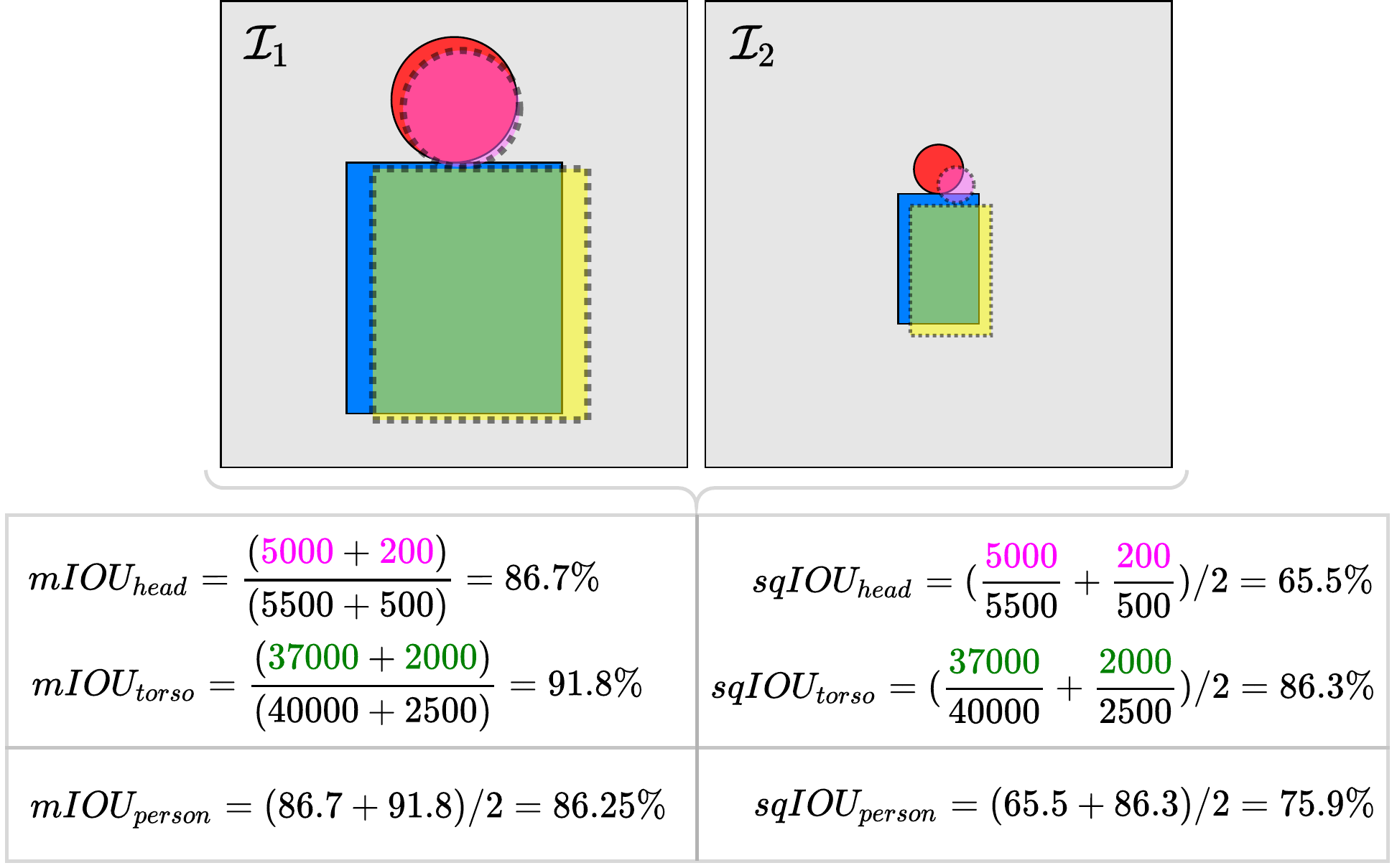}
  \caption{Toy example comparing mIOU and sqIOU with two images from \texttt{toy-person} category containing parts \texttt{head} and \texttt{torso} . `Red' and 'blue' represent  ground-truth, `pink' and 'green' represent prediction overlap areas. mIOU fails to reflect the bad segmentation of head in image $\mathcal{I}_2$ while sqIOU is fairer.}
  \label{fig:sqiou_eg}
\end{figure}

The calculation for sqIOU and sqAvg is similar to that of mIOU. Due to their formulation, mIOU and mAvg~\cite{gmnet,bsanet}
tend to be dominated by contributions from bigger\footnote{Informally, an instance is deemed ``big'' if it is among the largest instances for an object part category by area.} instances. In contrast, sqIOU and sqAvg weight parts of all sizes equally -- compare Eqn.~\ref{eqn:miou-p} and \ref{eqn:sqiou-p} and also see the toy example in Fig.~\ref{fig:sqiou_eg}. Therefore, sqIOU and sqAvg can be considered a more `fair' measure for segmentation quality.

\section{Experimental Results}
\label{sec:experiments}

For evaluation, we compare the performance of \FLOATsys\ with BSANet~\cite{bsanet}, GMNet~\cite{gmnet} and CO-Rank~\cite{co-rank}. As a baseline, we train a DeepLab-v3~\cite{chen2017deeplab} model with independently paired object category and associated part names (e.g. \texttt{cow left eye}, \texttt{cow right ear}) as labels. BSANet and CO-Rank report results on Pascal-Part-58 while GMNet additionally reports results on Pascal-Part-108. We report results on all variants of the Pascal-Part dataset, including our newly introduced Pascal-Part-201. To enable comparison, we train GMNet and BSANet on our dataset, Pascal-Part-201. For evaluation, we employ the mIOU, mAvg and sqIOU, sqAvg metrics described previously (Sec.~\ref{sec:metrics}). In addition, we analyze the relative contribution of various components in \FLOATsys\ via ablation studies. Full results table can be found in {\color{ProcessBlue}Appendix F}.

\subsection{Pascal-Part-201}

Table~\ref{tab:results201} shows the category-wise and overall performance on Pascal-Part-201. Overall, we see that \FLOATsys\ outperforms baselines and existing approaches by a significantly large margin. We obtain large gains of 10.8\% on mIOU and 8.1\% on sqIOU relative to the baseline. We outperform the next best method BSANet~\cite{bsanet} by large margins of 8.6\% on mIOU and 7.5\% on sqIOU as well. 

Empirically, we obtain significant sqIOU gains of 10\%-30\% on small parts -- for e.g. \texttt{left/right eye}, \texttt{left/right ear}, \texttt{left/right horn} etc. of `animate' categories such as \texttt{bird}, \texttt{cat}, \texttt{cow}. For `inanimate' categories (e.g. \texttt{bus}, \texttt{car}, \texttt{aeroplane}), we obtain sqIOU improvements in the range of 5\%-11\% on small parts such as \texttt{front/back plate}, \texttt{left/right wing}. The performance improvement is also similarly substantial for most parts containing side components (`left/right' or `front/back').

\subsection{Pascal-Part-58 and Pascal-Part-108}

\begin{table}
  \centering
  \begin{tabular}{@{}lccccc@{}}
    \toprule
    \footnotesize{Method} & \footnotesize{Dataset} & \footnotesize{mIOU} & \footnotesize{mAvg} & \footnotesize{sqIOU} & \footnotesize{sqAvg} \\
    \midrule
    \footnotesize{Baseline} & \multirow{5}{*}[-2pt]{\ts{58}} & \ts{54.3} & \ts{55.4} & \ts{46.0} & \ts{48.4} \\
    \footnotesize{BSANet\cite{bsanet}} && \ts{58.2} & \ts{58.9} & \ts{49.3} & \ts{51.5} \\
    \footnotesize{GMNet\cite{gmnet}} && \ts{59.0} & \ts{61.8} & \ts{49.4} & \ts{54.3} \\
    \footnotesize{CO-Rank\cite{co-rank}} && \ts{60.7} & \ts{60.6} & \ts{-} & \ts{-}\\
    \footnotesize{\textbf{FLOAT}} && \tsb{61.0} & \tsb{64.2} & \tsb{54.2} & \tsb{57.1} \\
    \midrule
    \footnotesize{Baseline} & \multirow{4}{*}[-2pt]{\ts{108}} & \ts{41.3} & \ts{43.6} & \ts{32.2} & \ts{36.1} \\
    \footnotesize{BSANet\cite{bsanet}} && \ts{45.9} & \ts{48.4} & \ts{36.6} & \ts{41.0} \\
    \footnotesize{GMNet\cite{gmnet}} && \ts{45.8} & \ts{50.5} & \ts{35.8} & \ts{41.9} \\
    \footnotesize{\textbf{FLOAT}} && \tsb{48.0} & \tsb{53.0} & \tsb{40.5} & \tsb{45.6} \\
    \bottomrule
  \end{tabular}
  \caption{Results on Pascal-Part-58, Pascal-Part-108: FLOAT outperforms the baseline and other existing methods on mIOU and with a significant gap on sqIOU. Missing CO-Rank entries are due to incomplete official codebase and missing details in the paper.}
  \label{tab:results58_108}
\end{table}

\begin{figure*}
  \centering
    \includegraphics[width=\linewidth]{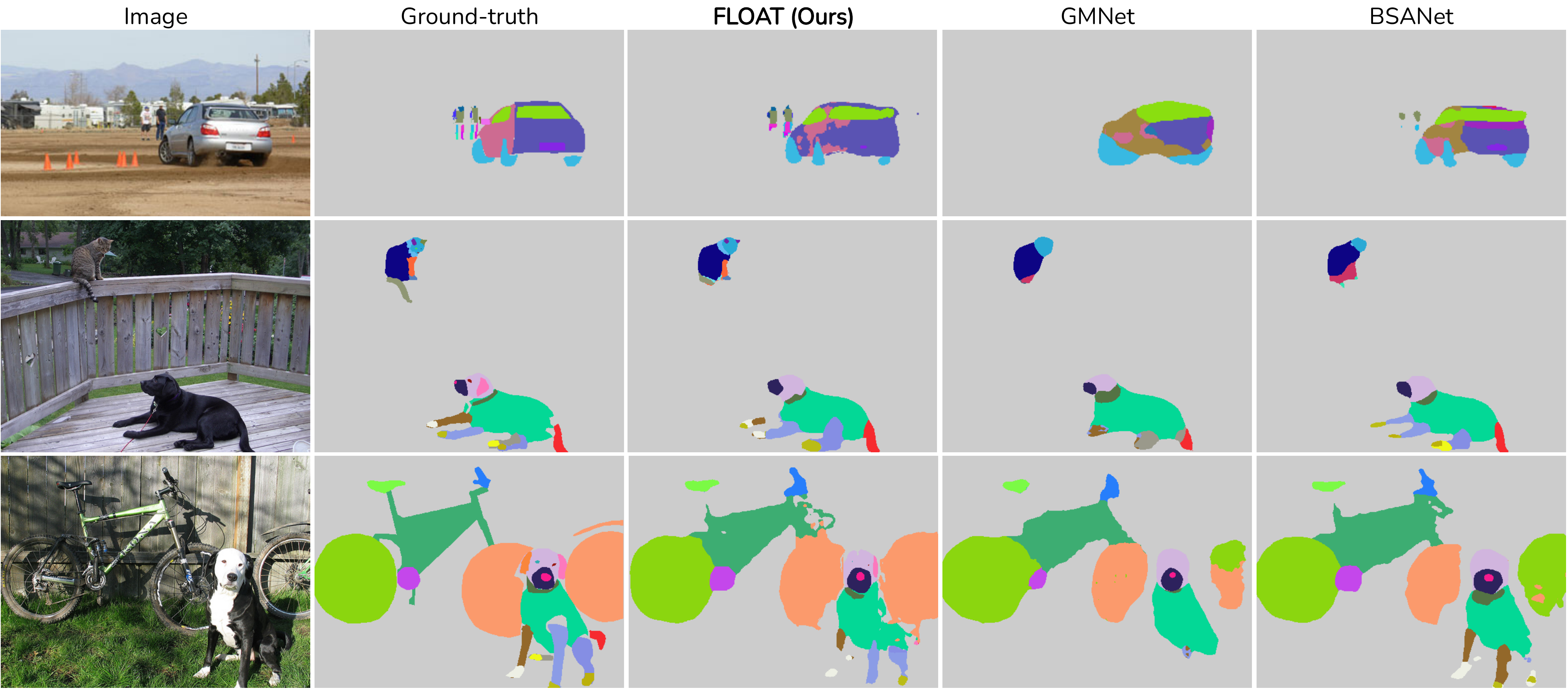}
    \caption{Qualitative comparison on Pascal-Part-201. We observe that FLOAT gets small objects parts -- \texttt{person} in the upper image, \texttt{cat} in the middle image. FLOAT also gets the left-right and front-back correct -- \texttt{leg(s)} of \texttt{dog} and \texttt{cat}, \texttt{side} of \texttt{car}, \texttt{wheel} of \texttt{bike}.}
    \label{fig:quali-comp}
    \vspace{-3mm}
\end{figure*}

\begin{table}[!t]
\setlength\tabcolsep{3pt}
  \centering
  \begin{tabular}{@{}l|c|c|cccccc|cc@{}}
    \toprule
    \footnotesize{Method} & \rotateT{Dataset} & \rotateT{Output Heads} & \rotateT{No Factorization}& \rotateT{Object} & \rotateT{Part} & \rotateT{Anim/Inanim} & \rotateT{Side} & \rotateT{Inference} \rotateT{Augmentation} & \ts{\textbf{mIOU}} & \ts{\textbf{sqIOU}} \\
    \midrule
    \footnotesize{Baseline} & \multirow{4}{*}[-2pt] {\ts{58}} & \ts{58} & \checkmark &&&&\ts{-}&& \ts{54.3} & \ts{46.0} \\
    \footnotesize{$\mathcal{M}_{obj} + \mathcal{M}_{part}$} && \ts{45} && \checkmark & \checkmark &&\ts{-}&& \ts{60.7} & \ts{51.5} \\
    \footnotesize{$\mathcal{F}$} && \ts{45} &&\checkmark&&\checkmark&\ts{-}&& \ts{60.9} & \ts{51.7} \\
    \footnotesize{\textbf{FLOAT}}  && \ts{45} &&\checkmark&&\checkmark&\ts{-}&\ts{IZR}& \tsb{61.0} & \tsb{54.2} \\
    \midrule
    \footnotesize{Baseline} & \multirow{4}{*}[-2pt]{\ts{108}} & \ts{108} &\checkmark &&&&\ts{-}&& \ts{41.3} & \ts{32.2} \\
    \footnotesize{$\mathcal{M}_{obj} + \mathcal{M}_{part}$} && \ts{68} && \checkmark & \checkmark &&\ts{-}&& \ts{46.1} & \ts{36.7} \\
    \footnotesize{$\mathcal{F}$} && \ts{68} &&\checkmark&&\checkmark&\ts{-}&& \ts{47.8} & \ts{38.4} \\
    \footnotesize{\textbf{FLOAT}} && \ts{68} &&\checkmark&&\checkmark&\ts{-}&\ts{IZR}& \tsb{48.0} & \tsb{40.5} \\
    \midrule
    \footnotesize{Baseline} &\multirow{7}{*}[-2pt]{\ts{201}} & \ts{201} & \checkmark &&&&&& \ts{26.3} & \ts{21.5} \\
    \footnotesize{$\mathcal{M}_{obj} + \mathcal{M}_{part}$} && \ts{119} && \checkmark & \checkmark &&&&\ts{29.1} & \ts{22.8} \\
    \footnotesize{$\mathcal{F}-D_{side}$} && \ts{119} &&\checkmark&&\checkmark&&& \ts{31.3} & \ts{24.1} \\
    \footnotesize{$\mathcal{F}$} && \ts{80} &&\checkmark&&\checkmark&\checkmark&& \ts{36.9} & \ts{27.8} \\
    \footnotesize{$\mathcal{F}$*} && \ts{80} &&\checkmark&&\checkmark&\checkmark*&& \ts{36.9} & \ts{27.6} \\
    \footnotesize{$\mathcal{F}$ + RCZ} && \ts{80} &&\checkmark&&\checkmark&\checkmark&\ts{RCZ}& \ts{36.6} & \ts{28.0} \\
    \footnotesize{\textbf{FLOAT}} && \ts{80} &&\checkmark&&\checkmark&\checkmark&\ts{IZR}& \tsb{37.1} & \tsb{29.6} \\
    \bottomrule
  \end{tabular}
  \caption{Ablation study: Starting from baseline with no factorization at all, we see that systematically adding components of FLOAT pipeline noticeably improves segmentation quality. $\mathcal{M}_{part}$ is combined decoder for all part-level labels, \textbf{FLOAT} $ = \mathcal{F} + IZR$ (see Fig.~\ref{fig:float}) is the proposed model. RCZ stands for Random Crop Zoom (see Sec.~\ref{sec:ablation}). The * indicates separate decoders for `left/right' and `front/back'. `Output heads' -- total number of output channels of a model. `No factorization' -- parts are labelled with concatenated category and associated part name. `Object' -- predicting object labels separately.}
  \label{tab:ablations}
\end{table}

We also show results on previously proposed datasets Pascal-Part-58~\cite{bsanet} and Pascal-Part-108~\cite{gmnet}. As shown in Table~\ref{tab:results58_108}, \FLOATsys\ framework achieves the best performance on both these datasets. In terms of mIOU, we outperform CO-Rank~\cite{co-rank} by 0.3\% on Pascal-Part-58 and GMNet~\cite{gmnet} by 2.0\%. In terms of sqIOU, we outperform other methods by large margins as well -- 4.8\% over GMNet and 4.9\% over BSANet. A similar trend is seen for Pascal-Part-108 with large improvements of 2.1\% on mIOU and 3.9\% on sqIOU over the next best method BSANet~\cite{bsanet}.

Overall, the results across existing and challenging new variants of Pascal-Part dataset demonstrate the strengths of our factorized label space setup. In particular, the increasing gains with increasing dataset complexity demonstrates the superior scaling capacity of the \FLOATsys\ framework. 

\subsection{Ablation Studies}
\label{sec:ablation}

We perform multiple experiments with ablative variant models of \FLOATsys\ to verify the effectiveness of our design choices. From the results in Table~\ref{tab:ablations}, we see that starting from baseline (first row in each dataset variant), systematically adding components of \FLOATsys\ pipeline noticeably improves segmentation quality. The gains are most apparent for Pascal-Part-201 dataset, particularly when factorized components are included. From the last two rows, we also see that IZR is a superior choice compared to Random Crop Zoom (RCZ) - a variant which uses random crops whose cardinality matches the number of objects in the scene. Some part names  in the original Pascal-Part dataset~\cite{chen2014detect} contain the side component `upper/lower'. We attempted to train a FLOAT variant with these components as outputs of $D_{side}$ decoder. However, the model failed to converge. We hypothesize  this is due to the drastically smaller quantum of training data compared to other \textit{side} attributes, i.e. `left/right' and `front/back'.

\subsection{Qualitative Analysis}

Fig.~\ref{fig:quali-comp} shows qualitative comparisons of our framework with existing approaches on Pascal-Part-201, reflecting the improvements gains we observe for mIOU and sqIOU metrics (Table~\ref{tab:results201}). FLOAT is visually superior at segmenting smaller object parts -- notice the significantly improved segmentation for parts in object categories \texttt{person} ( first row) and \texttt{cat} (second row). From the examples, we see that FLOAT is also better at learning directionality (`left/right', `front/back'). Similar improvements are evident from the examples provided in Figure \ref{fig:fig1} ({\color{ProcessBlue}Appendix E} contains additional examples). Some limitations of FLOAT include missing predictions for the smallest of parts (e.g. \texttt{eye} in people far from camera) and partial predictions for thin parts leading to disconnections.

\section{Conclusion}
\label{sec:conclusion}

FLOAT is a simple but effective framework for improving semantic segmentation performance in multi-object multi-part parsing. Our idea of \textit{factorized label space} is a key contribution which fully takes advantage of label-level intra/inter ontological relationships among objects and parts. The factorization not only enables scalability in terms of both object categories and part labels, but also improves segmentation performance substantially. Another key contribution is our inference-time zoom. By focusing only on object-centric regions of interest, IZR efficiently enhances segmentation quality without requiring explicit object feature guidance or other modifications to the part network setup. Apart from our framework, we introduce a new variant of Pascal-Part called Pascal-Part-201 which constitutes the most challenging benchmark dataset for the problem. Our experimental evaluation, using fairer versions of existing measures, shows that FLOAT clearly outperforms existing state-of-the-art approaches for existing and newly introduced Pascal-Part variants. The gains from our framework increase with increased part and object dataset complexity, empirically supporting our assertion of FLOAT's scalability. Although presented in a 2D scene parsing setting, we expect ideas from FLOAT to be useful for the 3D scene parsing counterpart and in general, for scenarios with appropriately factorizable attributes.
{\small
\bibliographystyle{ieee_fullname}
\bibliography{egbib}
}

\appendix

\title{Appendix - FLOAT: Factorized Learning of Object Attributes for Improved Multi-object Multi-part Scene Parsing}
\author{Rishubh Singh\textsuperscript{1} \hspace{1cm} Pranav Gupta\textsuperscript{2} \hspace{1cm} Pradeep Shenoy\textsuperscript{1} \hspace{1cm} Ravikiran Sarvadevabhatla\textsuperscript{2}\\
\textsuperscript{1}Google Research \hspace{1cm} \textsuperscript{2} IIIT Hyderabad\\
{\tt\small \{rishubh,shenoypradeep\}@google.com, \{ravi.kiran@,pranav.gu@research.\}iiit.ac.in}}
\maketitle

\textit{In this document we show some further  experimental results, animate/inanimate object group split, computational stats, visual results and used algorithms in details. We report mIOU and sqIOU for each part class for all three datatset(Pascal-Part-58,108 and 201)}

\section{Algorithm details}

\subsection{Top Down Merge}
The flowchart in the following page describes the “Top Down Merge” algorithm per pixel to obtain the final
label for that pixel (aggregation across the image gives the final prediction). As described in the paper, each
label consists of an object, a root part component and side component(s). For FLOAT, these are determined
separately and merged to obtain the final label at each pixel. For each pixel :

\begin{enumerate}

\item We obtain the object category predicted. We now have an “object” label.

\item Choose the part from the animate part map or the inanimate part map depending on the object category. We now have an “object part” label.

\item We now add side components :

\begin{enumerate}
\item Animate:
\begin{enumerate}
\item For animate categories, a part can have both left/right and front/back labels.
\item Depending on what side components the “object part” needs to match the original label
space, the same are added from the $S_{lr}$ and $S_{fb}$ side maps.
\item To make sure each pixel has a left/right and a front/back label, while taking the softmax, we ignore the background category prediction.
\end{enumerate}

\item Inanimate:
\begin{enumerate}
\item For animate categories, a part can have only one of left/right/front/back labels.

\begin{figure}[!t]
  \centering
\includegraphics[width=\linewidth]{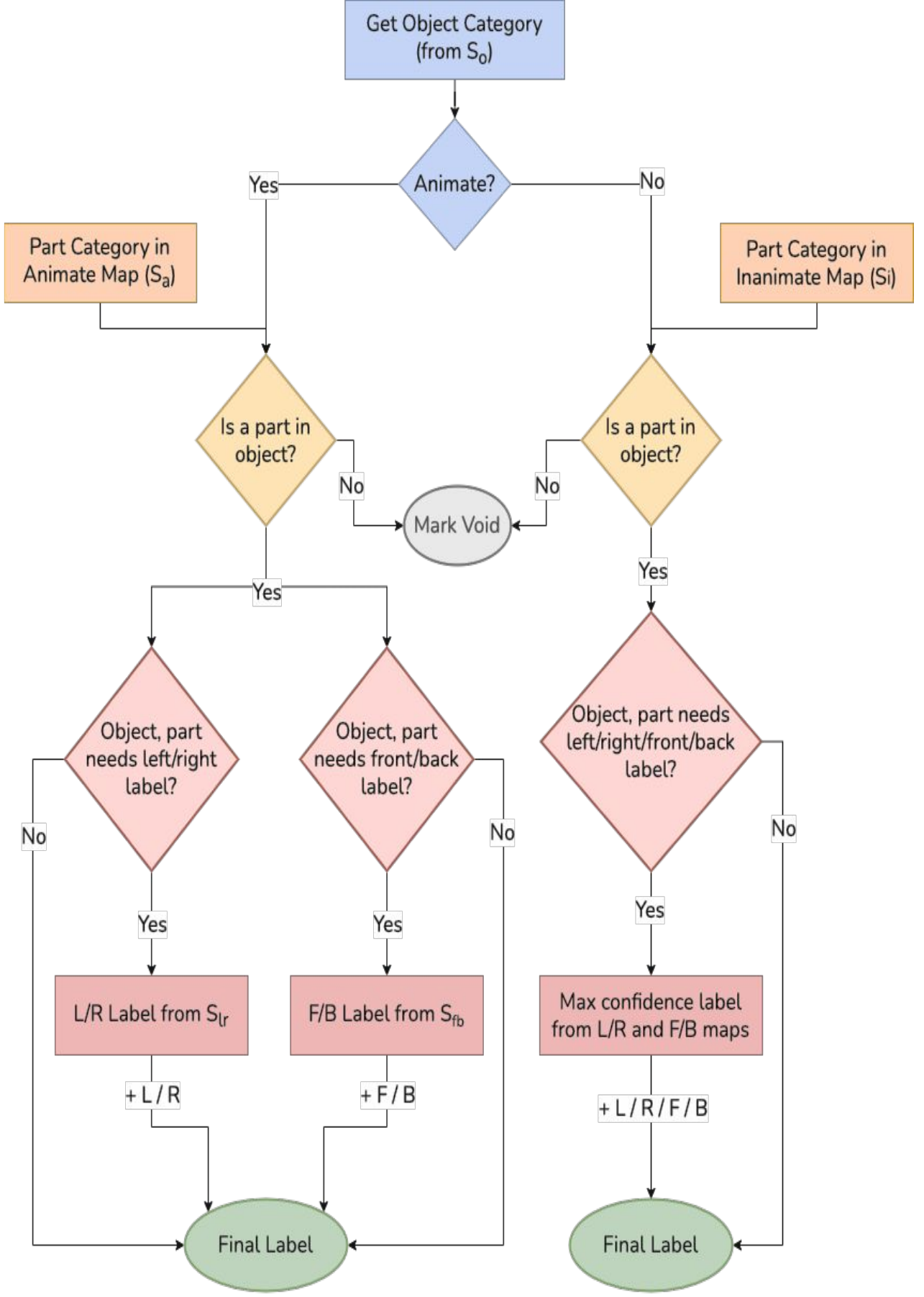}
\caption{}

\end{figure}

\begin{figure*}[!t]
  \centering
    \includegraphics[width=\linewidth]{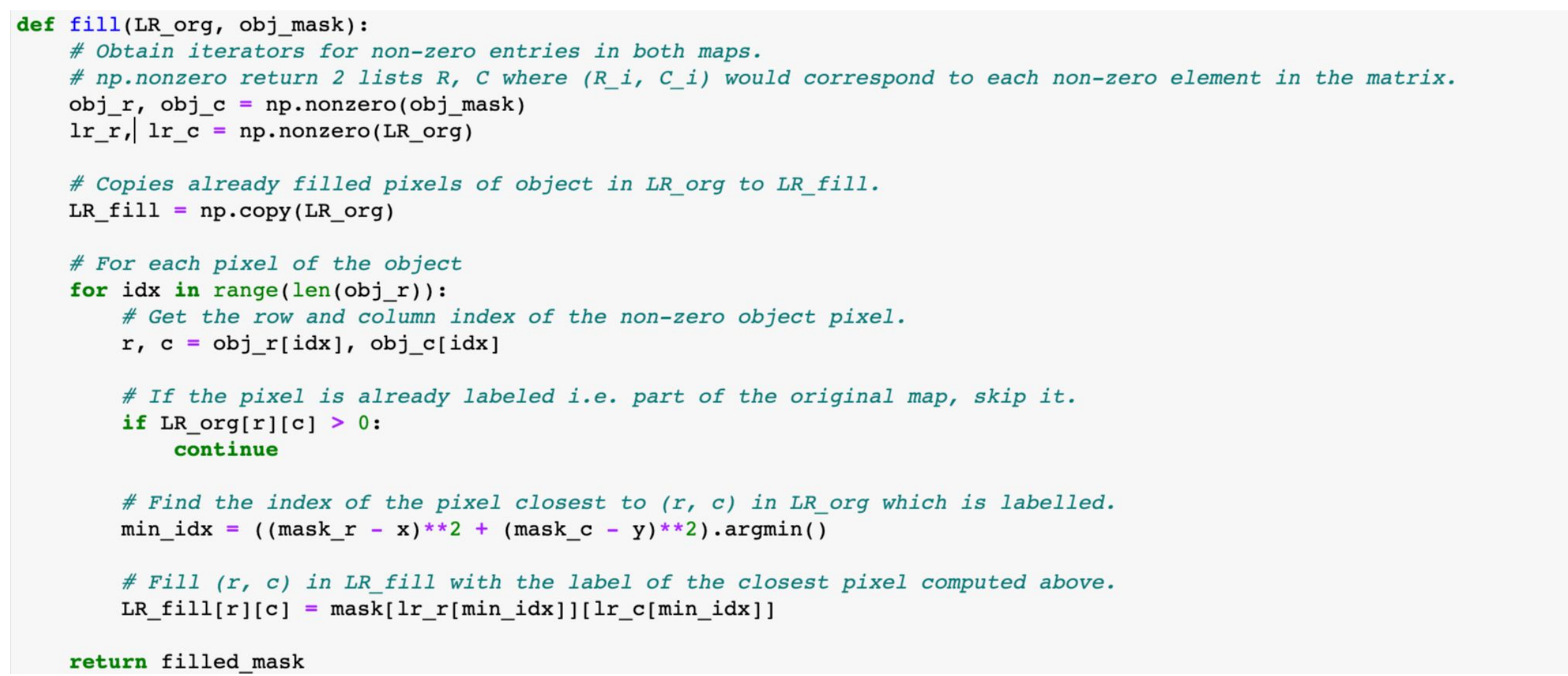}
    \captionsetup{justification=centering,margin=8cm}
        \caption{}
        \label{fig:flood_fill}
\end{figure*} 

\item We compute the combined Left-Right-Front-Back (LRFB) map by combining the Left-Right ($S_{lr}$) and Front-Back ($S_{fb}$) maps using confidence (softmax) values.
\item If the “object part” needs the side component, the same is added from the LRFB map.

\end{enumerate}
\end{enumerate}

\end{enumerate}
Hence, we get all components required from predicting the final label for each pixel :
{\bf “Object L/R F/B Part”} for animate and {\bf “Object L/R/F/B Part”} for inanimate objects.


\subsection{Flood Fill for Side Component Ground Truths}
As an approximation to a breadth-first search style flood fill for generation ground truths, we compute the side component label for each pixel by allocating it the same label as the one closest to it in the map without flood fill.
\\
\\
Let’s assume, for an object, the original left-right map is LR\_org and the map we want to compute is LR\_fill.
The 0-1 object mask for the under consideration object is obj\_mask. The python snippet for computing LR\_fill
given LR\_org and obj\_mask is given in Figure \ref{fig:flood_fill} (FB\_fill can be computed from FB\_org using the same):

\subsection{Illustration of Factorization described in Introduction}

\begin{figure}[h]
  \centering
\includegraphics[width=\linewidth]{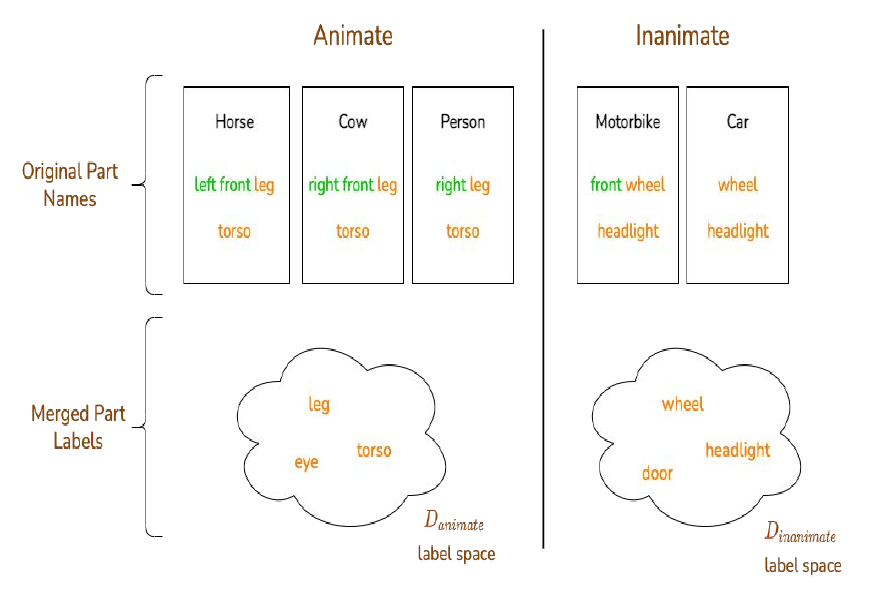}
    \caption{}
    \label{fig:factorization}
\end{figure}

Objects are split into animate and inanimate groups. The parts in each group share root components which
are merged to form the label set for part prediction for each set of objects. See Figure \ref{fig:factorization} for pictorial illustration.

\section{Animate/Inanimate object group split}
\begin{figure}[h]
\includegraphics[width=\linewidth]{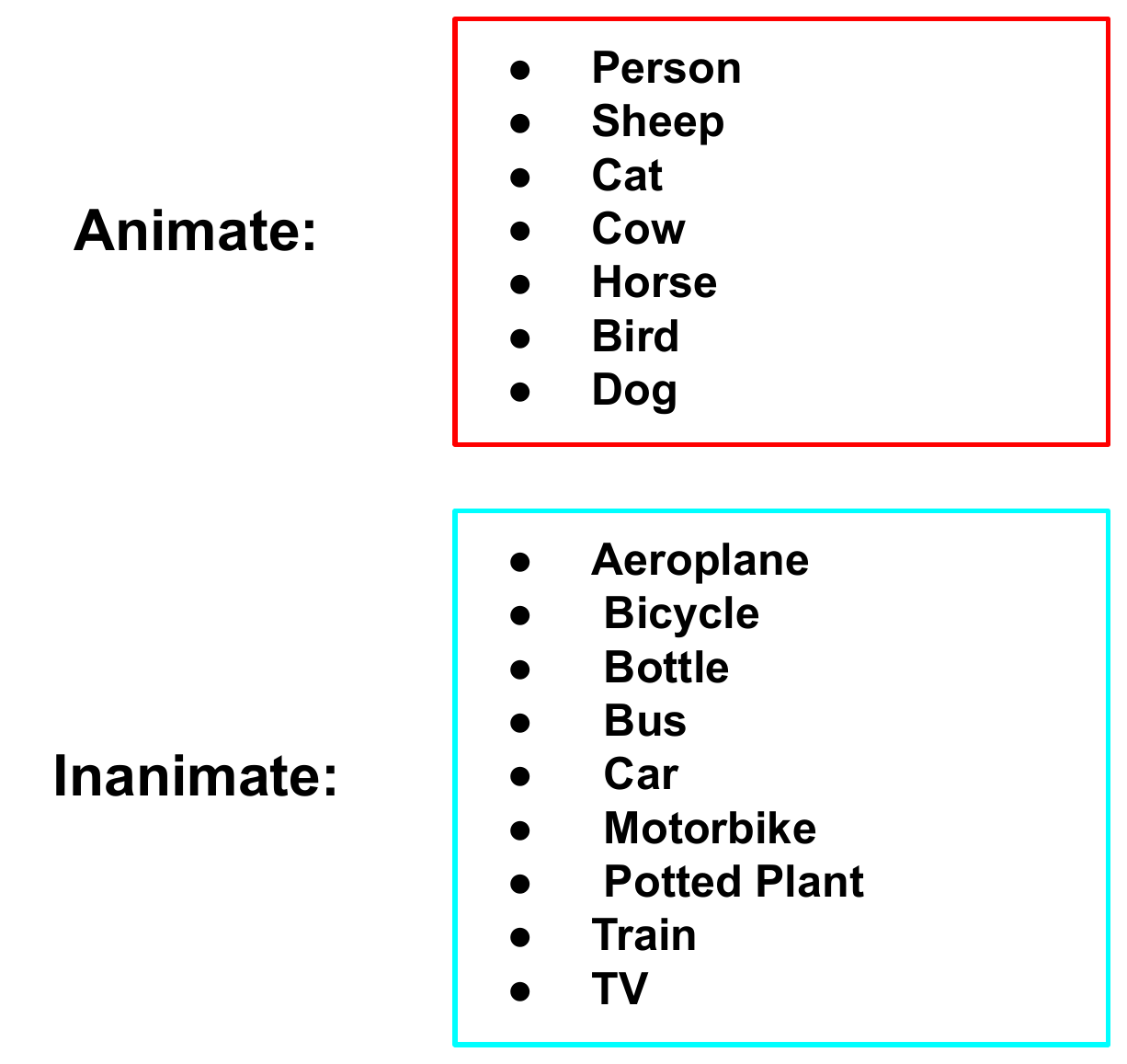}
    \caption{}
    \label{fig:group_split}
\end{figure}

There are total 7 animate and 10 inanimate object categories with parts. See Figure \ref{fig:group_split} for group split.

\section{Memory and Compute:}
Table~\ref{tab:table} summarizes the compute requirements for various models and datasets configurations. Despite the somewhat larger number of parameters compared to other models, FLOAT trains faster and provides significant segmentation performance gains.

\begin{table}[h]
  \centering
  \setlength\tabcolsep{3.2pt}
  \renewcommand{\arraystretch}{0.7}
  \begin{tabular}{@{}lccccc@{}}
    \toprule
    \footnotesize{Method} & \footnotesize{Dataset} & \footnotesize{Params(M)} & \footnotesize{Train Time (mins)} & \footnotesize{Test Time (secs)} \\
    \midrule
    \footnotesize{BSANet} & \multirow{3}{*}[-2pt]{\ts{58}} & \tsb{63.9} & \ts{40.2} & \tsb{0.45} \\
    \footnotesize{GMNet} && \ts{124.9} & \ts{33.6} & \ts{0.49} \\
    \footnotesize{\textbf{FLOAT} (45)} && \ts{135.4} & \tsb{30.1} & \ts{1.02 (0.55)} \\
    \midrule
    \footnotesize{BSANet} & \multirow{3}{*}[-2pt]{\ts{108}} & \tsb{63.9} & \ts{43.7} & \tsb{0.72} \\
    \footnotesize{GMNet} && \ts{124.9} & \ts{37.2} & \ts{0.75} \\
    \footnotesize{\textbf{FLOAT} (68)} && \ts{135.4} & \tsb{34.8} & \ts{1.38 (0.80)} \\
    \midrule
    \footnotesize{BSANet} & \multirow{3}{*}[-2pt]{\ts{201}} & \tsb{64.0} & \ts{47.1} & \tsb{1.30} \\
    \footnotesize{GMNet} && \ts{124.9} & \ts{40.3} & \ts{1.35} \\
    \footnotesize{\textbf{FLOAT} (80)} && \ts{153.6} & \tsb{38.6} & \ts{2.14 (1.43)} \\
    \bottomrule
  \end{tabular}
  \caption{Compute comparisons of FLOAT with previous methods. Train time is per epoch. Test time is per instance. (Batch size $=$ 5). Total output heads for FLOAT given in brackets under method. Test time in brackets for FLOAT quotes time without IZR.}
  \label{tab:table}
\end{table}

\section{Limitations}

\begin{itemize}

\item Partial predictions of objects with only a few parts visible in the scene.

\item Bad predictions around complicated boundaries, eg - rider on a bicycle.

\item Missing some very small/obscure objects in an image.

\item Missing some predictions for objects with bad lighting / extremely varying shapes.

\end{itemize}

\clearpage

\section{Results:}
\subsection{Part-58}
\begin{figure}[h]
  \centering
    \includegraphics[width=\textwidth]{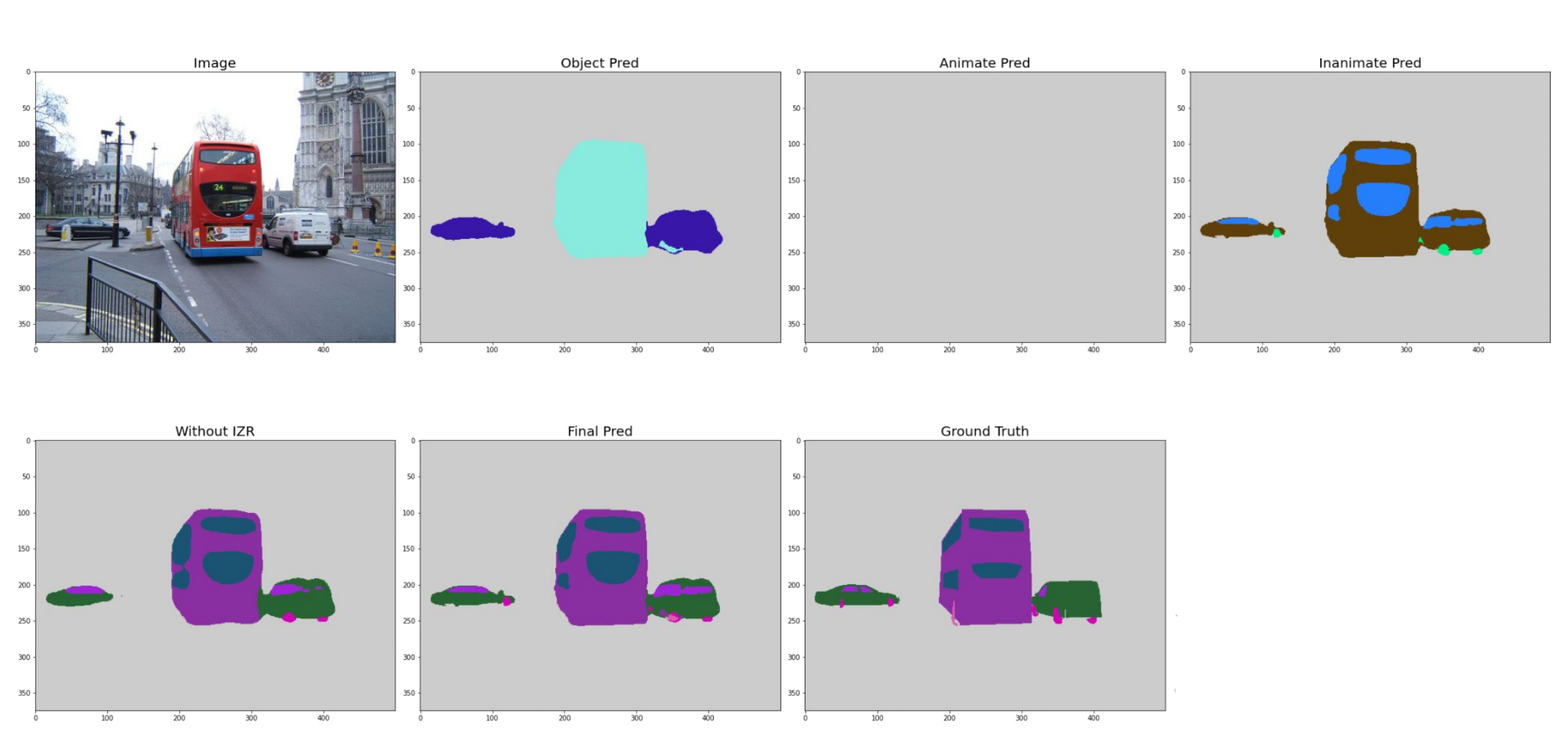}
    \captionsetup{justification=centering,margin=8cm}
        \caption{}
\end{figure}   

\begin{figure}[h]
  \centering
    \includegraphics[width=\textwidth]{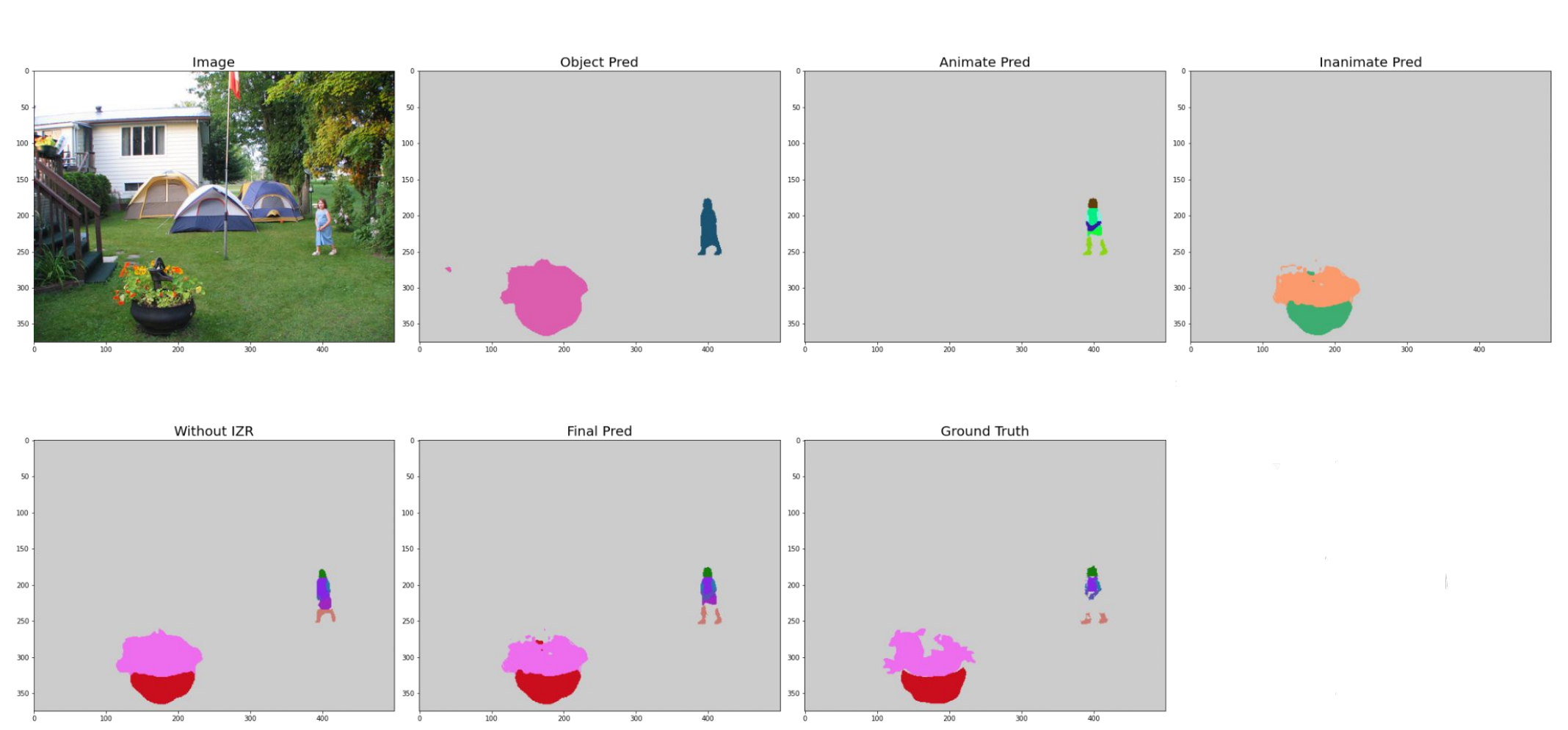}
    \captionsetup{justification=centering,margin=8cm}
        \caption{}
\end{figure} 

\clearpage

\subsection{Part-108}
\begin{figure}[h]
  \centering
    \includegraphics[width=\textwidth]{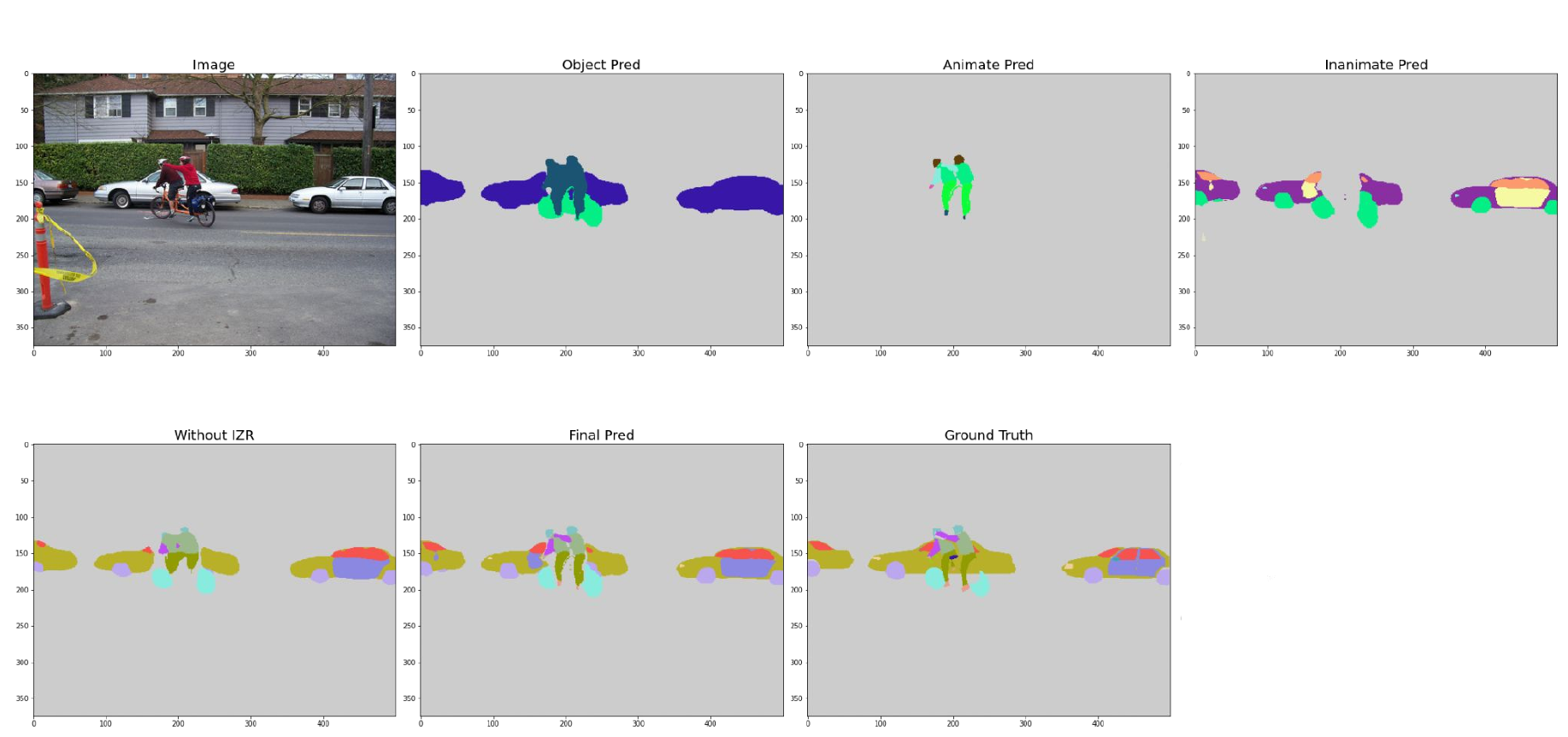}
    \captionsetup{justification=centering,margin=8cm}
        \caption{}
\end{figure} 

\begin{figure}[h]
  \centering
    \includegraphics[width=\textwidth]{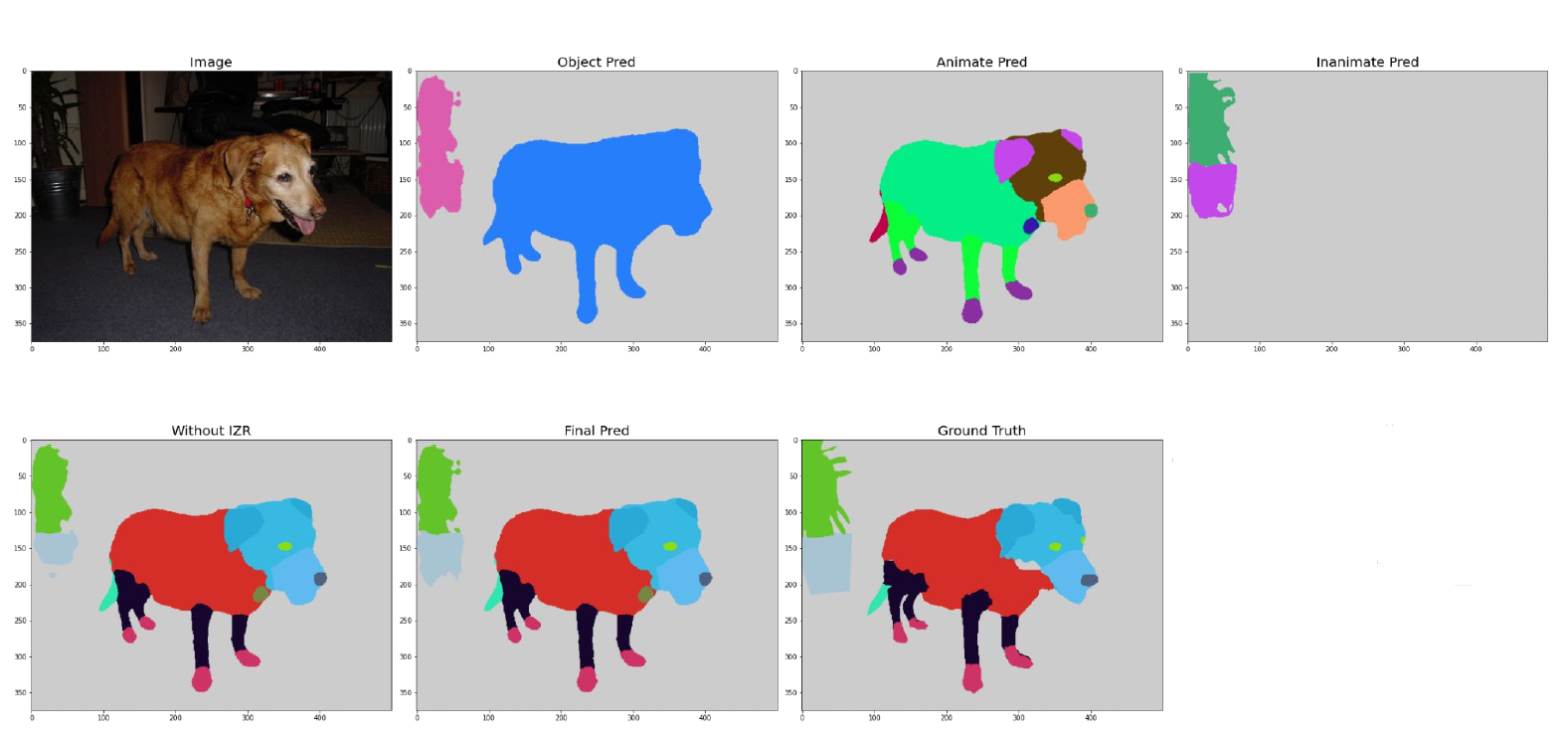}
    \captionsetup{justification=centering,margin=8cm}
        \caption{}
\end{figure} 

\clearpage

\subsection{Part-201}
\begin{figure}[h]
  \centering
    \includegraphics[width=\textwidth, height=9.5cm]{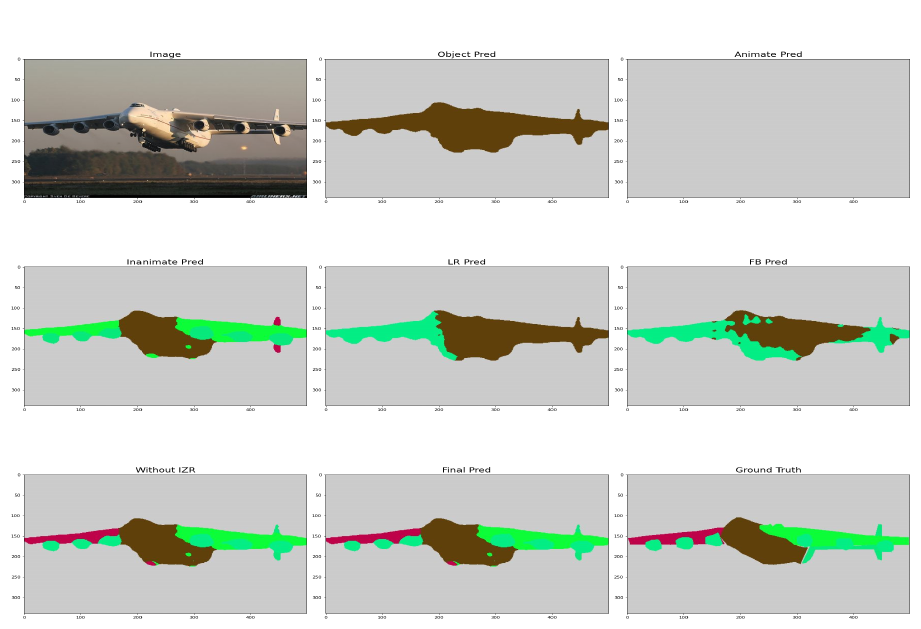}
    \captionsetup{justification=centering,margin=8cm}
        \caption{}
\end{figure}   

\begin{figure}[h]
  \centering
    \includegraphics[width=\textwidth, height=9.5cm]{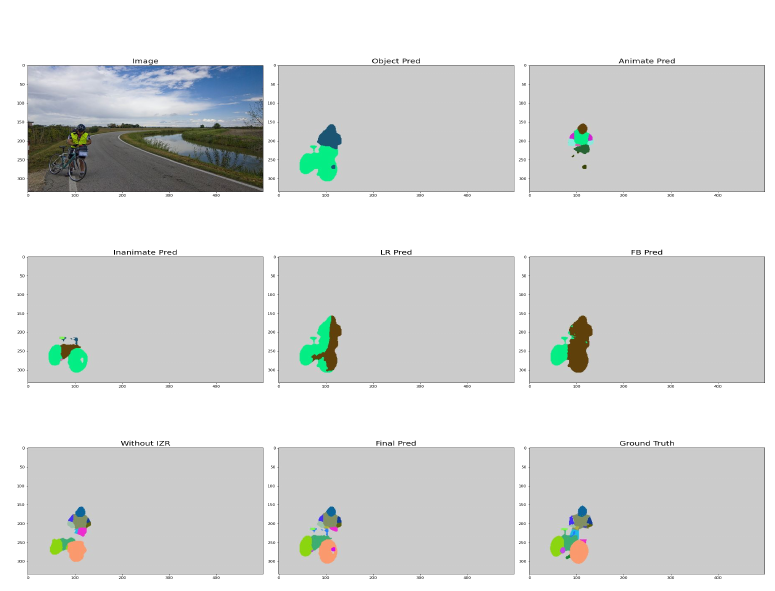}
    \captionsetup{justification=centering,margin=8cm}
        \caption{}
\end{figure} 

\clearpage

\section{Pascal-Part Results}

\subsection{Pascal-Part-58  mIOU comparison}

\begin{table}[h]
  \centering
\begin{tabular}{@{}|l|cccc|@{}}
    \toprule
    \footnotesize{Part} & \ts{Baseline} & \ts{BSANet} & \ts{GMNet} & \tsb{FLOAT} \\
    \midrule 
\footnotesize{Background} & \ts{90.1} & \ts{91.6} & \ts{92.7} & \tsb{92.9} \\ 
\footnotesize{Aeroplane Body} & \ts{65.3} & \ts{69.5} & \ts{69.6} & \tsb{70.2} \\ 
\footnotesize{Aeroplane Engine} & \ts{24.9} & \ts{28.9} & \ts{25.7} & \tsb{29.7} \\ 
\footnotesize{Aeroplane Wing} & \ts{33.9} & \ts{37.3} & \ts{34.2} & \tsb{37.6} \\ 
\footnotesize{Aeroplane Stern} & \ts{56.3} & \ts{57.2} & \ts{57.2} & \tsb{58.3} \\ 
\footnotesize{Aeroplane Wheel} & \ts{43.8} & \tsb{51.8} & \ts{46.8} & \ts{45.7} \\ 
\footnotesize{Bicycle Wheel} & \ts{77.8} & \ts{76.9} & \tsb{81.3} & \ts{81.2} \\
\bottomrule
  \end{tabular}
  \label{tab:results_201_iou_1}
\end{table}
\begin{table}[H]
  \centering
  \begin{tabular}{@{}|l|cccc|@{}}
    \toprule
    \footnotesize{Part} & \ts{Baseline} & \ts{BSANet} & \ts{GMNet} & \tsb{FLOAT} \\
\midrule
\footnotesize{Person Lower Leg} & \ts{38.6} & \ts{42.0} & \ts{40.2} & \tsb{42.4} \\ 
\footnotesize{Person Upper Leg} & \ts{43.8} & \ts{46.3} & \ts{44.5} & \tsb{47.4} \\ 
\footnotesize{Potted Plant Pot} & \ts{47.3} & \ts{51.3} & \tsb{56.0} & \ts{50.8} \\ 
\footnotesize{Potted Plant Plant} & \ts{52.4} & \ts{55.5} & \ts{56.4} & \tsb{58.9} \\ 
\footnotesize{Sheep Head} & \ts{60.9} & \ts{63.6} & \tsb{70.8} & \ts{70.6} \\ 
\footnotesize{Sheep Leg} & \ts{8.6} & \ts{19.4} & \ts{14.3} & \tsb{24.4} \\ 
\footnotesize{Sheep Torso} & \ts{68.3} & \ts{71.7} & \ts{75.6} & \tsb{76.0} \\ 
\footnotesize{Sofa} & \ts{43.2} & \ts{42.6} & \ts{56.1} & \tsb{69.1} \\ 
\footnotesize{Train} & \ts{76.6} & \ts{80.9} & \ts{85.0} & \tsb{86.0} \\ 
\footnotesize{TV Screen} & \ts{69.5} & \ts{72.3} & \tsb{77.0} & \ts{72.0} \\ 
\footnotesize{TV Frame} & \ts{44.4} & \ts{49.0} & \tsb{54.1} & \ts{47.3} \\ 
    \bottomrule
  \end{tabular}
  \label{tab:results_58_iou_2}
\end{table}

\subsection{Pascal-Part-58 sqIOU comparison}

\begin{table}[h]
  \centering
  \begin{tabular}{@{}|l|cccc|@{}}
    \toprule
    \footnotesize{Part} & \ts{Baseline} & \ts{BSANet} & \ts{GMNet} & \tsb{FLOAT} \\
    
\midrule 
\footnotesize{Background} & \ts{89.6} & \ts{90.2} & \ts{91.0} & \tsb{91.2} \\ 
\footnotesize{Aeroplane Body} & \ts{60.9} & \ts{63.9} & \ts{62.2} & \tsb{64.2} \\ 
\footnotesize{Aeroplane Engine} & \ts{22.9} & \tsb{39.4} & \ts{34.1} & \ts{38.6} \\ 
\footnotesize{Aeroplane Wing} & \ts{30.8} & \tsb{35.5} & \ts{30.2} & \ts{35.1} \\ 
\footnotesize{Aeroplane Stern} & \ts{48.2} & \ts{49.2} & \ts{48.9} & \tsb{50.5} \\ 
\footnotesize{Aeroplane Wheel} & \ts{30.5} & \ts{32.7} & \ts{27.7} & \tsb{39.2} \\ 
\footnotesize{Bicycle Wheel} & \ts{68.7} & \ts{66.8} & \ts{68.4} & \tsb{72.9} \\
\bottomrule
  \end{tabular}
  \label{tab:results_201_iou_1}
\end{table}
\begin{table}[H]
  \centering
  \begin{tabular}{@{}|l|cccc|@{}}
    \toprule
    \footnotesize{Part} & \ts{Baseline} & \ts{BSANet} & \ts{GMNet} & \tsb{FLOAT} \\
\midrule
\footnotesize{Dog Leg} & \ts{42.5} & \ts{50.3} & \ts{47.5} & \tsb{52.9} \\ 
\footnotesize{Dog Tail} & \ts{22.7} & \ts{25.5} & \ts{27.7} & \tsb{36.2} \\ 
\footnotesize{Dog Torso} & \ts{58.3} & \ts{62.3} & \ts{62.2} & \tsb{63.9} \\ 

\footnotesize{Horse Head} & \ts{61.3} & \ts{63.7} & \ts{63.9} & \tsb{70.7} \\ 
\footnotesize{Horse Tail} & \ts{30.8} & \ts{33.6} & \ts{34.1} & \tsb{39.7} \\ 
\footnotesize{Horse Leg} & \ts{47.7} & \ts{52.4} & \ts{50.7} & \tsb{56.3} \\ 
\footnotesize{Horse Torso} & \ts{68.5} & \ts{70.0} & \ts{71.9} & \tsb{74.2} \\
\bottomrule
  \end{tabular}
  \label{tab:results_58_sqiou_2}
\end{table}

\subsection{Pascal-Part-108 mIOU comparison}

\begin{table}[h]
  \centering
  \begin{tabular}{@{}|l|cccc|@{}}
    \toprule
    \footnotesize{Part} & \ts{Baseline} & \ts{BSANet} & \ts{GMNet} & \tsb{FLOAT} \\
\midrule 
\footnotesize{Background} & \ts{90.2} & \ts{91.8} & \ts{92.7} & \tsb{92.9} \\ 
\footnotesize{Aeroplane Body} & \ts{60.9} & \ts{69.6} & \ts{61.9} & \tsb{70.0} \\ 
\footnotesize{Aeroplane Engine} & \ts{53.2} & \ts{56.2} & \ts{27.2} & \tsb{57.6} \\ 
\footnotesize{Aeroplane Wing} & \ts{27.9} & \tsb{37.2} & \ts{34.3} & \ts{34.3} \\ 
\footnotesize{Aeroplane Stern} & \ts{24.7} & \ts{29.4} & \tsb{57.4} & \ts{27.6} \\ 
\footnotesize{Aeroplane Wheel} & \ts{40.9} & \ts{51.0} & \tsb{51.5} & \ts{45.5} \\ 
\footnotesize{Bicycle Wheel} & \ts{76.4} & \ts{77.1} & \ts{80.2} & \tsb{81.2} \\ 
\footnotesize{Bicycle Saddle} & \ts{34.1} & \tsb{38.3} & \ts{38.0} & \ts{36.9} \\ 
\footnotesize{Bicycle Handlebar} & \ts{23.3} & \tsb{25.2} & \ts{22.4} & \ts{18.8} \\ 
\footnotesize{Bicycle Chainwheel} & \ts{42.3} & \ts{41.6} & \ts{44.1} & \tsb{53.6} \\ 
\footnotesize{Bird Head} & \ts{51.5} & \ts{66.6} & \ts{65.3} & \tsb{70.0} \\ 
\footnotesize{Bird Beak} & \ts{40.4} & \ts{51.3} & \ts{44.3} & \tsb{56.0} \\ 
\footnotesize{Bird Torso} & \ts{61.7} & \tsb{67.3} & \ts{64.8} & \ts{63.4} \\ 
\footnotesize{Bird Neck} & \ts{27.5} & \tsb{34.5} & \ts{28.4} & \ts{34.3} \\ 
\footnotesize{Bird Wing} & \ts{35.9} & \tsb{41.3} & \ts{37.2} & \ts{36.2} \\ 
\footnotesize{Bird Leg} & \ts{23.5} & \tsb{30.8} & \ts{23.8} & \ts{25.5} \\ 
\footnotesize{Bird Foot} & \ts{13.9} & \tsb{18.3} & \ts{17.7} & \ts{17.4} \\ 
\footnotesize{Bird Tail} & \ts{28.1} & \tsb{35.7} & \ts{32.5} & \ts{33.6} \\ 
\footnotesize{Boat} & \ts{53.7} & \ts{60.7} & \ts{69.2} & \tsb{74.8} \\ 
\footnotesize{Bottle Cap} & \ts{30.4} & \ts{31.0} & \ts{33.4} & \tsb{37.6} \\ 
\footnotesize{Bottle Body} & \ts{63.7} & \ts{71.4} & \tsb{78.7} & \ts{72.4} \\ 
\bottomrule
  \end{tabular}
  \label{tab:results_201_iou_1}
\end{table}
\begin{table}[H]
  \centering
  \begin{tabular}{@{}|l|cccc|@{}}
    \toprule
    \footnotesize{Part} & \ts{Baseline} & \ts{BSANet} & \ts{GMNet} & \tsb{FLOAT} \\
\midrule
\footnotesize{Bus Side} & \ts{70.1} & \ts{74.9} & \ts{75.7} & \tsb{77.1} \\ 
\footnotesize{Bus Roof} & \ts{7.5} & \ts{6.3} & \ts{13.5} & \tsb{17.1} \\ 
\footnotesize{Bus Mirror} & \ts{2.1} & \tsb{8.0} & \ts{6.6} & \ts{0.0} \\ 
\footnotesize{Bus Plate} & \ts{0.0} & \ts{0.0} & \ts{0.0} & \ts{0.0} \\ 
\footnotesize{Bus Door} & \ts{40.1} & \tsb{47.4} & \ts{38.1} & \ts{40.6} \\ 
\footnotesize{Bus Wheel} & \ts{53.8} & \ts{56.6} & \ts{56.7} & \tsb{59.8} \\ 
\footnotesize{Bus Headlight} & \ts{25.6} & \ts{31.0} & \ts{30.4} & \tsb{44.7} \\
\bottomrule
  \end{tabular}
  \label{tab:results_201_iou_1}
\end{table}
\begin{table}[H]
  \centering
  \begin{tabular}{@{}|l|cccc|@{}}
    \toprule
    \footnotesize{Part} & \ts{Baseline} & \ts{BSANet} & \ts{GMNet} & \tsb{FLOAT} \\
\midrule
\footnotesize{Motorbike Wheel} & \ts{69.6} & \ts{71.6} & \ts{73.6} & \tsb{76.4} \\ 
\footnotesize{Motorbike Hbar} & \ts{0.0} & \ts{0.0} & \ts{0.0} & \ts{0.0} \\ 
\footnotesize{Motorbike Saddle} & \ts{0.0} & \ts{0.0} & \tsb{0.8} & \ts{0.1} \\ 
\footnotesize{Motorbike Hlight} & \ts{25.8} & \ts{21.3} & \ts{28.5} & \tsb{35.0} \\ 
\footnotesize{Person Head} & \ts{68.2} & \ts{71.3} & \ts{69.3} & \tsb{72.6} \\ 
\footnotesize{Person Eye} & \ts{35.1} & \ts{44.6} & \ts{38.7} & \tsb{49.6} \\ 
\footnotesize{Person Ear} & \ts{37.4} & \ts{46.4} & \ts{41.4} & \tsb{47.6} \\
\bottomrule
  \end{tabular}
  \label{tab:results_108_iou_3}
\end{table}

\subsection{Pascal-Part-108 sqIOU comparison}

\begin{table}[h]
  \centering
  \begin{tabular}{@{}|l|cccc|@{}}
    \toprule
    \footnotesize{Part} & \ts{Baseline} & \ts{BSANet} & \ts{GMNet} & \tsb{FLOAT} \\
\midrule 
\footnotesize{Background} & \ts{88.7} & \ts{90.6} & \ts{91.2} & \tsb{91.3} \\ 
\footnotesize{Aeroplane Body} & \ts{55.1} & \ts{63.9} & \ts{62.3} & \tsb{64.5} \\ 
\footnotesize{Aeroplane Engine} & \ts{47.1} & \ts{49.1} & \ts{32.5} & \tsb{50.2} \\ 
\footnotesize{Aeroplane Wing} & \ts{22.9} & \tsb{35.5} & \ts{30.0} & \ts{31.0} \\ 
\footnotesize{Aeroplane Stern} & \ts{29.7} & \ts{40.2} & \tsb{50.1} & \ts{34.5} \\ 
\footnotesize{Aeroplane Wheel} & \ts{25.9} & \ts{33.2} & \ts{29.4} & \tsb{35.1} \\ 
\footnotesize{Bicycle Wheel} & \ts{63.1} & \ts{67.3} & \ts{68.4} & \tsb{71.2} \\ 
\footnotesize{Bicycle Saddle} & \ts{23.4} & \ts{27.0} & \ts{26.0} & \tsb{28.2} \\ 

\footnotesize{Bicycle Handlebar} & \ts{15.7} & \tsb{18.9} & \ts{16.3} & \ts{16.7} \\ 
\bottomrule
  \end{tabular}
  \label{tab:results_201_iou_1}
\end{table}
\begin{table}[H]
  \centering
  \begin{tabular}{@{}|l|cccc|@{}}
    \toprule
    \footnotesize{Part} & \ts{Baseline} & \ts{BSANet} & \ts{GMNet} & \tsb{FLOAT} \\
\midrule
\footnotesize{Bicycle Chainwheel} & \ts{29.4} & \ts{30.2} & \ts{29.1} & \tsb{33.5} \\ 

\footnotesize{Bird Head} & \ts{42.7} & \ts{47.6} & \ts{45.8} & \tsb{52.4} \\ 

\footnotesize{Bird Beak} & \ts{22.8} & \ts{30.8} & \ts{25.9} & \tsb{37.5} \\ 
\footnotesize{Bird Torso} & \ts{49.6} & \tsb{57.0} & \ts{54.2} & \ts{54.7} \\ 
\footnotesize{Bird Neck} & \ts{23.7} & \ts{30.2} & \ts{26.0} & \tsb{33.1} \\ 
\footnotesize{Bird Wing} & \ts{31.2} & \ts{31.5} & \ts{33.2} & \tsb{36.4} \\ 
\footnotesize{Bird Leg} & \ts{8.2} & \tsb{14.6} & \ts{10.0} & \ts{14.3} \\
\bottomrule
  \end{tabular}
  \label{tab:results_201_iou_1}
\end{table}
\begin{table}[H]
  \centering
  \begin{tabular}{@{}|l|cccc|@{}}
    \toprule
    \footnotesize{Part} & \ts{Baseline} & \ts{BSANet} & \ts{GMNet} & \tsb{FLOAT} \\
\midrule
\footnotesize{Dog Leg} & \ts{33.6} & \ts{40.0} & \ts{37.5} & \tsb{41.9} \\ 
\footnotesize{Dog Paw} & \ts{32.8} & \ts{39.8} & \ts{35.9} & \tsb{41.7} \\ 
\footnotesize{Dog Tail} & \ts{22.6} & \ts{28.1} & \ts{27.4} & \tsb{33.2} \\ 
\footnotesize{Dog Muzzle} & \ts{53.8} & \ts{60.5} & \ts{57.8} & \tsb{62.7} \\ 
\footnotesize{Horse Head} & \ts{46.6} & \ts{49.9} & \ts{49.3} & \tsb{54.2} \\ 
\footnotesize{Horse Ear} & \ts{29.3} & \ts{38.2} & \ts{35.1} & \tsb{44.4} \\ 
\footnotesize{Horse Muzzle} & \ts{54.1} & \ts{54.7} & \ts{55.6} & \tsb{61.6} \\
\bottomrule
  \end{tabular}
  \label{tab:results_108_sqiou_3}
\end{table}

\subsection{Pascal-Part-201 mIOU comparison}

\begin{table}[H]
  \centering
  \begin{tabular}{@{}|l|cccc|@{}}
    \toprule
    \footnotesize{Part} & \ts{Baseline} & \ts{BSANet} & \ts{GMNet} & \tsb{FLOAT} \\
    \midrule 
\footnotesize{Background } & \ts{91.0} & \ts{91.2} & \ts{90.7} & \tsb{92.5} \\ 
\footnotesize{Aeroplane Body} & \ts{67.3} & \tsb{71.1} & \ts{62.2} & \ts{68.9} \\ 
\footnotesize{Aeroplane Engine} & \ts{27.0} & \tsb{30.0} & \ts{19.2} & \ts{28.5} \\ 
\footnotesize{Aeroplane Left Wing} & \ts{3.8} & \ts{10.7} & \ts{7.2} & \tsb{28.6} \\ 
\footnotesize{Aeroplane Right Wing} & \ts{19.6} & \ts{20.3} & \ts{13.4} & \tsb{25.6} \\ 
\footnotesize{Aeroplane Stern} & \ts{53.3} & \tsb{56.7} & \ts{48.1} & \ts{55.3} \\ 
\footnotesize{Aeroplane Tail} & \ts{0.0} & \ts{0.0} & \ts{0.0} & \ts{0.0} \\
\bottomrule
  \end{tabular}
  \label{tab:results_201_iou_1}
\end{table}
\begin{table}[H]
  \centering
  \begin{tabular}{@{}|l|cccc|@{}}
    \toprule
    \footnotesize{Part} & \ts{Baseline} & \ts{BSANet} & \ts{GMNet} & \tsb{FLOAT} \\
\midrule
\footnotesize{Car Headlight} & \tsb{54.4} & \ts{51.5} & \ts{39.4} & \ts{54.3} \\ 
\footnotesize{Car Left Mirror} & \ts{12.6} & \ts{14.0} & \ts{1.4} & \tsb{21.0} \\ 
\footnotesize{Car Left Side} & \ts{20.5} & \ts{20.5} & \ts{16.6} & \tsb{28.7} \\ 
\footnotesize{Car Right Mirror} & \ts{0.3} & \ts{7.1} & \ts{0.0} & \tsb{17.8} \\ 
\footnotesize{Car Right Side} & \ts{16.7} & \ts{18.7} & \ts{14.6} & \tsb{29.0} \\ 
\footnotesize{Car Roof} & \ts{17.4} & \tsb{27.8} & \ts{11.7} & \ts{22.5} \\ 
\footnotesize{Car Wheel} & \ts{68.3} & \ts{69.2} & \ts{63.7} & \tsb{72.4} \\
\bottomrule
  \end{tabular}
  \label{tab:results_201_iou_1}
\end{table}
\begin{table}[H]
  \centering
  \begin{tabular}{@{}|l|cccc|@{}}
    \toprule
    \footnotesize{Part} & \ts{Baseline} & \ts{BSANet} & \ts{GMNet} & \tsb{FLOAT} \\
\midrule
\footnotesize{Dog Left Front Leg} & \ts{9.4} & \ts{18.6} & \ts{14.2} & \tsb{32.3} \\ 
\footnotesize{Dog Left Front Paw} & \ts{8.2} & \ts{16.6} & \ts{16.2} & \tsb{30.2} \\ 
\footnotesize{Dog Muzzle} & \ts{67.7} & \tsb{70.3} & \ts{63.2} & \ts{64.9} \\ 
\footnotesize{Dog Neck} & \tsb{27.5} & \ts{27.4} & \ts{20.1} & \ts{25.6} \\ 
\footnotesize{Dog Nose} & \ts{64.0} & \tsb{69.2} & \ts{55.1} & \ts{66.0} \\ 
\footnotesize{Dog Right Back Leg} & \ts{18.0} & \ts{12.3} & \ts{13.5} & \tsb{24.4} \\ 
\footnotesize{Dog Right Back Paw} & \ts{2.5} & \ts{5.5} & \ts{2.5} & \tsb{18.1} \\
\bottomrule
  \end{tabular}
  \label{tab:results_201_iou_1}
\end{table}
\begin{table}[H]
  \centering
  \begin{tabular}{@{}|l|cccc|@{}}
    \toprule
    \footnotesize{Part} & \ts{Baseline} & \ts{BSANet} & \ts{GMNet} & \tsb{FLOAT} \\
\midrule
\footnotesize{Person Mouth} & \ts{52.2} & \ts{53.6} & \ts{30.2} & \tsb{57.9} \\ 
\footnotesize{Person Neck} & \ts{51.3} & \ts{52.2} & \ts{42.6} & \tsb{52.9} \\ 
\footnotesize{Person Nose} & \ts{59.2} & \ts{58.2} & \ts{40.1} & \tsb{61.1} \\ 
\footnotesize{Person Right Ear} & \ts{16.3} & \ts{19.8} & \ts{9.8} & \tsb{47.8} \\ 
\footnotesize{Person Right Eye} & \ts{22.8} & \ts{17.9} & \ts{13.7} & \tsb{47.3} \\ 
\footnotesize{Person Right Eyebrow} & \ts{0.0} & \ts{3.6} & \ts{0.4} & \tsb{12.2} \\ 
\footnotesize{Person Right Foot} & \ts{9.3} & \ts{12.4} & \ts{7.2} & \tsb{18.4} \\
\bottomrule
  \end{tabular}
  \label{tab:results_201_iou_4}
\end{table}

\subsection{Pascal-Part-201 sqIOU comparison}

\begin{table}[H]
  \centering
  \begin{tabular}{@{}|l|cccc|@{}}
    \toprule
    \footnotesize{Part} & \ts{Baseline} & \ts{BSANet} & \ts{GMNet} & \tsb{FLOAT} \\
    
\midrule 
\footnotesize{Background } & \ts{89.6} & \ts{89.9} & \ts{89.4} & \tsb{90.8} \\ 
\footnotesize{Aeroplane Body} & \ts{61.3} & \tsb{65.0} & \ts{54.2} & \ts{62.2} \\ 
\footnotesize{Aeroplane Engine} & \ts{36.4} & \tsb{40.6} & \ts{19.1} & \ts{38.6} \\ 
\footnotesize{Aeroplane Left Wing} & \ts{3.8} & \ts{10.6} & \ts{3.7} & \tsb{22.2} \\ 
\footnotesize{Aeroplane Right Wing} & \ts{19.6} & \ts{16.1} & \ts{9.1} & \tsb{22.7} \\ 
\footnotesize{Aeroplane Stern} & \ts{47.5} & \tsb{49.4} & \ts{39.7} & \ts{46.4} \\ 
\footnotesize{Aeroplane Tail} & \ts{0.0} & \ts{0.0} & \ts{0.0} & \ts{0.0} \\
\bottomrule
  \end{tabular}
  \label{tab:results_201_iou_1}
\end{table}
\begin{table}[H]
  \centering
  \begin{tabular}{@{}|l|cccc|@{}}
    \toprule
    \footnotesize{Part} & \ts{Baseline} & \ts{BSANet} & \ts{GMNet} & \tsb{FLOAT} \\
\midrule
\footnotesize{Car Headlight} & \ts{30.3} & \ts{30.0} & \ts{13.5} & \tsb{31.7} \\ 

\footnotesize{Car Left Mirror} & \ts{4.9} & \ts{7.5} & \ts{0.0} & \tsb{7.7} \\ 
\footnotesize{Car Left Side} & \ts{19.2} & \ts{18.9} & \ts{15.1} & \tsb{24.6} \\ 
\footnotesize{Car Right Mirror} & \ts{0.2} & \ts{5.0} & \ts{0.0} & \tsb{8.1} \\ 
\footnotesize{Car Right Side} & \ts{17.3} & \ts{17.6} & \ts{10.3} & \tsb{25.9} \\ 
\footnotesize{Car Roof} & \ts{17.3} & \ts{20.1} & \ts{6.8} & \tsb{22.9} \\ 
\footnotesize{Car Wheel} & \ts{50.7} & \ts{49.2} & \ts{38.9} & \tsb{56.6} \\
\bottomrule
  \end{tabular}
  \label{tab:results_201_iou_1}
\end{table}
\begin{table}[H]
  \centering
  \begin{tabular}{@{}|l|cccc|@{}}
    \toprule
    \footnotesize{Part} & \ts{Baseline} & \ts{BSANet} & \ts{GMNet} & \tsb{FLOAT} \\
\midrule
\footnotesize{Dog Left Front Leg} & \ts{7.0} & \ts{14.6} & \ts{9.5} & \tsb{28.6} \\ 
\footnotesize{Dog Left Front Paw} & \ts{8.9} & \ts{15.5} & \ts{11.8} & \tsb{28.8} \\ 
\footnotesize{Dog Muzzle} & \ts{61.5} & \ts{60.7} & \ts{49.6} & \tsb{62.4} \\ 
\footnotesize{Dog Neck} & \tsb{21.8} & \ts{20.2} & \ts{8.2} & \ts{21.3} \\ 
\footnotesize{Dog Nose} & \ts{44.6} & \ts{45.0} & \ts{24.7} & \tsb{48.0} \\ 
\footnotesize{Dog Right Back Leg} & \ts{12.2} & \ts{8.3} & \ts{7.9} & \tsb{17.8} \\ 
\footnotesize{Dog Right Back Paw} & \ts{2.5} & \ts{5.6} & \ts{0.3} & \tsb{17.3} \\
\bottomrule
  \end{tabular}
  \label{tab:results_201_iou_1}
\end{table}
\begin{table}[H]
  \centering
  \begin{tabular}{@{}|l|cccc|@{}}
    \toprule
    \footnotesize{Part} & \ts{Baseline} & \ts{BSANet} & \ts{GMNet} & \tsb{FLOAT} \\
\midrule
\footnotesize{Person Mouth} & \ts{25.3} & \ts{25.4} & \ts{10.5} & \tsb{31.9} \\ 
\footnotesize{Person Neck} & \ts{39.7} & \ts{35.7} & \ts{26.7} & \tsb{42.3} \\ 
\footnotesize{Person Nose} & \ts{30.5} & \ts{27.2} & \ts{14.4} & \tsb{36.2} \\ 
\footnotesize{Person Right Ear} & \ts{9.1} & \ts{9.9} & \ts{1.8} & \tsb{23.3} \\ 
\footnotesize{Person Right Eye} & \ts{5.7} & \ts{3.9} & \ts{5.6} & \tsb{19.6} \\ 
\footnotesize{Person Right Eyebrow} & \ts{0.0} & \ts{0.8} & \ts{0.0} & \tsb{3.1} \\ 
\footnotesize{Person Right Foot} & \ts{6.7} & \ts{6.9} & \ts{7.1} & \tsb{16.3} \\
\bottomrule
  \end{tabular}
 
  \label{tab:results_201_sqiou_4}
\end{table}

\end{document}